\documentclass{article}

\PassOptionsToPackage{numbers}{natbib}

\usepackage[preprint]{neurips_2024}

\usepackage[utf8]{inputenc} 
\usepackage[T1]{fontenc}    
\usepackage{hyperref}       
\usepackage{url}            
\usepackage{booktabs}       
\usepackage{amsfonts}       
\usepackage{nicefrac}       
\usepackage{microtype}      
\usepackage{xcolor}         

\usepackage{amsmath}
\usepackage{graphicx}
\usepackage[capitalize]{cleveref}
\setcitestyle{square,numbers}
\hypersetup{colorlinks,linkcolor={blue},citecolor={green},urlcolor={magenta}}
\usepackage{arydshln}
\usepackage{marvosym}  

\def \ie {\emph{i.e.}~}
\def \eg {\emph{e.g.}~}

\def \vs {\emph{v.s.}~}

\definecolor{darkgreen}{rgb}{0.04,0.63,0.07}

\title{VLPrompt: Vision-Language Prompting for\\ Panoptic Scene Graph Generation}

\author{
  Zijian Zhou${}^{1}$ \quad 
  Miaojing Shi${}^{2}$\textsuperscript{\Letter} \quad
  Holger Caesar${}^{3}$ \quad \\
  ${}^{1}$Department of Informatics, King's College London\\
  ${}^{2}$College of Electronic and Information Engineering, Tongji University\\
  ${}^{3}$Intelligent Vehicles Lab, Delft University of Technology\\
}

\begin{document}

\maketitle
\footnotetext{\textsuperscript{\Letter} Corresponding author.}

\begin{abstract}
Panoptic Scene Graph Generation (PSG) aims at achieving a comprehensive image understanding by simultaneously segmenting objects and predicting relations among objects.
However, the long-tail problem among relations leads to unsatisfactory results in real-world applications.
Prior methods predominantly rely on vision information or utilize limited language information, such as object or relation names, thereby overlooking the utility of language information.
Leveraging the recent progress in Large Language Models (LLMs), we propose to use language information to assist relation prediction, particularly for rare relations.
To this end, we propose the \textbf{V}ision-\textbf{L}anguage \textbf{Prompt}ing (\textbf{VLPrompt}) model, which acquires vision information from images and language information from LLMs.
Then, through a prompter network based on attention mechanism, it achieves precise relation prediction.
Our extensive experiments show that VLPrompt significantly outperforms previous state-of-the-art methods on the PSG dataset, proving the effectiveness of incorporating language information and alleviating the long-tail problem of relations.
Code is available at \url{https://github.com/franciszzj/VLPrompt}.
\end{abstract}
\section{Introduction}
\label{sec:introduction}

Panoptic Scene Graph Generation (PSG)~\cite{yang2022panoptic} extends Scene Graph Generation (SGG)~\cite{lu2016visual} by incorporating panoptic segmentation~\cite{kirillov2019panoptic} to capture richer and more detailed representations of images, including both ``thing''~\cite{lin2014microsoft} and ``stuff''~\cite{caesar2018coco} classes.
PSG constructs a directed graph to represent an image, where nodes signify objects 
and edges capture the relations between objects.
As a bridge between vision and language, PSG has a multitude of downstream applications such as visual question answering~\cite{hildebrandt2020scene}, image captioning~\cite{gao2018image, chen2020say}, and visual reasoning~\cite{aditya2018image, shi2019explainable}; furthermore, it can also benefit relevant fields like embodied navigation~\cite{singh2023scene} and robotic action planning~\cite{amiri2022reasoning}.

Notwithstanding, the current performance of PSG~\cite{yang2022panoptic, zhou2023hilo, wang2023pair, li2023panoptic} remains unsatisfactory, limiting its downstream applications.
The essential reason lies in the severe long-tail problem in relation categories: for instance, in the PSG dataset~\cite{yang2022panoptic}, the top three most frequent relation categories account for over 50\% of entire samples, with numerous rare relations appearing less than 1\%.
PSG models thus struggle to accurately predict these rare relations.

\begin{figure}[t]
    \centering
    \vspace{-2mm}
    \includegraphics[width=1.0\linewidth]{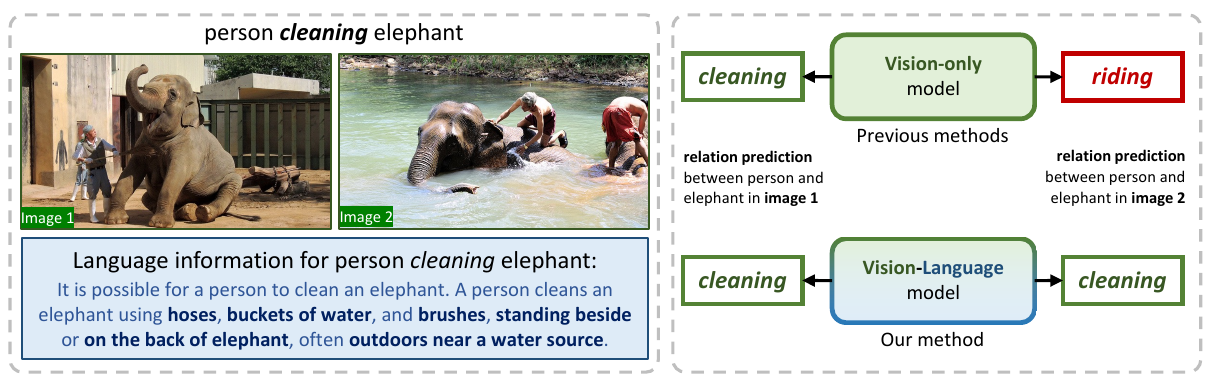}
    \vspace{-8mm}
    \caption{Comparison between previous PSG methods and ours.
    \textbf{Left}: Images of ``person \textit{cleaning} elephant'' in two different scenes, accompanied by snippets of descriptions about ``person \textit{cleaning} elephant'' obtained from LLMs.
    \textbf{Right}: Previous vision-only models can predict the \textit{cleaning} relation between the person and elephant in image 1, but often classify image 2's relation as \textit{riding} due to the person's position on the back of the elephant.
    Our vision-language model, enriched with language information, precisely identifies the \textit{cleaning} relation in both images.
    }
    \vspace{-4mm}
    \label{fig:motivation}
\end{figure}

Recent methods~\cite{yang2022panoptic, zhou2023hilo, wang2023pair}
have made progress in addressing the long-tail problem, mainly exploiting the strength of vision information for relation prediction, whilst overlooking the language information in PSG.
The integration of language information is however important to provide additional common sense
knowledge for objects and their relations. 
For example, in the two images of Fig.~\ref{fig:motivation}, the relation between the person and the elephant is \textit{cleaning}.
In the top right image, where the person is on the elephant's back \textit{cleaning} it, this scenario can easily lead previous vision-only models to classify the relation as \textit{riding}.
In contrast, our vision-language model can utilize language information like ``a person cleans an elephant using brushes on the back of the elephant'', thus precisely predicting the relation as \textit{cleaning}.

In SGG task, some methods~\cite{he2022towards, dong2022stacked} have recognized the importance of incorporating language information besides vision. 
However, the way language information is utilized in these works is limited to category names of objects or relations, providing no further context and hence not fully addressing the long-tail problem.
The same observation goes to methods like~\cite{gu2019scene, zareian2020bridging}, which integrate knowledge graphs into the SGG task.
With the rapid development of Large Language Models (LLMs)~\cite{chatgpt, touvron2023llama}, acquiring richer language information - instead of merely the concepts of objects or relations - becomes much easier than before.

In this paper, we introduce a novel \textbf{V}ision-\textbf{L}anguage \textbf{Prompt}ing (\textbf{VLPrompt}) model, which leverages rich language information from LLMs to help predict panoptic scene graphs in visual images.
The language information serves as a powerful supplement to relation prediction, especially for rare ones.
Our model comprises three parts.
The first is the \emph{vision feature extractor}, where we process the input image with a panoptic segmentation network adapted from Mask2Former \cite{cheng2022masked} to extract features of different objects.
We pair and concatenate these object features and integrate their corresponding spatial information to form the vision prompting features.
In contrast, in the second part, the \emph{language feature extractor}, we employ the chain-of-thought technique~\cite{wei2022chain} to design various prompts, aiming to stimulate LLMs to propose context-rich language information for potential relations between a subject-object pair or judge a specific subject-relation-object triplet. These two functions are realized via a carefully designed relation proposer prompt and relation judger prompt.  
Subsequently, these language descriptions are transformed into language features using a pre-trained text encoder.
Finally, in the third part, the \emph{vision-language prompter}, we design a novel dual attention-based prompter network to facilitate the interaction between vision features and the two complimentary language features respectively, resulting into two sets of relation predictions. They are combined via a MLP-based gating network to take the strength of both for final relation prediction. The whole VLPrompt is trained end-to-end.

To the best of our knowledge, we are the first to utilize language information generated by LLMs for the PSG task.
Extensive experiments on the PSG dataset~\cite{yang2022panoptic} demonstrate that our VLPrompt drastically enhances the PSG performance (\eg improving the R@100 from 43.0 to 52.4 and mR@100 from 33.1 to 53.7), underscoring the significance of integrating language information for PSG.


\vspace{-2mm}
\section{Related Work}
\label{sec:related_work}

\vspace{-2mm}
\subsection{Scene Graph Generation}
\label{sec:sgg}

Scene Graph Generation (SGG)~\cite{lu2016visual} is a crucial task in scene understanding and has garnered widespread attention in the computer vision community.
In recent years, numerous methods~\cite{xu2017scene, zellers2018neural, tang2019learning, lin2020gps, li2022sgtr, shit2022relationformer, zhang2022fine, yu2023visually} have achieved notable progress.
Various model architectures have been proposed, such as intricately designed message passing structures~\cite{li2017vip, dai2017detecting, li2017scene, zellers2018neural, gu2019scene, hu2022neural}, attention-based networks~\cite{zheng2019visual, qi2019attentive}, tree-based networks~\cite{zhang2017visual, hung2020contextual} and DETR-based networks~\cite{li2022sgtr, shit2022relationformer, cong2023reltr}.
Specifically, to address the long-tail problem, some methods enhance the prediction accuracy of rare relations through data re-sampling~\cite{li2021bipartite} and loss re-weighting~\cite{kang2023skew}.
Relevant techniques that have been developed include constructing enhanced datasets~\cite{zhang2022fine, yu2023visually}, grouping relations for training~\cite{dong2022stacked}, constructing multi-stage hierarchical training~\cite{deng2022hierarchical}, and designing de-bias loss functions~\cite{yu2020cogtree, kang2023skew}.
Most methods leverage images as sole inputs. Recently, some methods~\cite{lu2016visual, liao2019natural, zhang2019large, hwang2018tensorize, dupty2020visual} have begun exploring language information or knowledge graphs in SGG; specifically, the explored language information is so far confined to basic language concepts of objects or relations.

\vspace{-2mm}
\subsection{Panoptic Scene Graph Generation}
\label{sec:psg}

Panoptic Scene Graph Generation (PSG)~\cite{yang2022panoptic} has emerged as a novel task in scene understanding in recent years.
Unlike SGG~\cite{lu2016visual}, PSG employs panoptic segmentation instead of bounding boxes to represent objects, enabling a more comprehensive understanding.
Similar to SGG methods, current methods in PSG~\cite{yang2022panoptic, zhou2023hilo, wang2023pair, li2023panoptic} also mainly rely on the image input and do not utilize any language information.
For instance, PSGTR~\cite{yang2022panoptic} build a baseline PSG model by adding a relation prediction head to DETR~\cite{carion2020end}.
PSGFormer~\cite{yang2022panoptic} advances PSGTR by separately modeling objects and relations in two transformer decoders and introducing an interaction mechanism. 
Recently, HiLo~\cite{zhou2023hilo} addresses the long-tail problem by specializing different network branches in learning both high and low frequency relations. 
PairNet~\cite{wang2023pair} develops a novel framework using a pair proposal network to filter sparse pairwise relations, improving PSG performance.
Unlike these methods, we are the first to propose a PSG method, relying on both vision and language inputs.

\vspace{-2mm}
\subsection{Large Language Models for Vision Tasks}
\label{sec:llm}
LLMs have led to large improvements in natural language processing tasks~\cite{min2023recent}.
They are normally trained on extensive text corpora by learning to autoregressively predict the next word, hence encapsulate a broad spectrum of common sense knowledge of linguistic patterns, cultural norms, and basic worldly facts.
Prominent examples of LLMs include GPT series~\cite{chatgpt}, Llama series~\cite{touvron2023llama1, touvron2023llama}, Bard~\cite{bard}, and Claude~\cite{claude}, with Llama series being open-source publicly available.
Given the extensive common sense information contained in LLMs, some researchers start to propose multimodal sockets to LLMs and apply them to various vision tasks, such as recognition~\cite{huang2023inject, wang2023all}, detection~\cite{tang2023cotdet, zhang2023next}, segmentation~\cite{lai2023lisa, zhou2023text}, visual question answering~\cite{liu2023large, xenos2023simple}, image reasoning~\cite{chen2023large} and robotic navigation~\cite{tsai2023multimodal, shah2023lm}; nevertheless, there are no such models specifically designed for panoptic scene graph generation so far.
In contrast, our method designs various prompts to stimulate LLMs to elicit rich language information to enhance relation prediction.

\vspace{-2mm}
\section{Method}
\label{sec:method}
\vspace{-2mm}

In this section, we introduce our method, \emph{VLPrompt}.
Given an image ($\mathcal{I} \in \mathbb{R}^{H \times W \times 3}$) and language description ($\mathcal{T}$) generated from LLMs, we extract vision and language features from them to predict panoptic scene graph $\mathcal{G} = \{\mathcal{O}, \mathcal{R}\}$, \ie $\mathcal{G} = \text{VLPrompt} (\mathcal{I}, \mathcal{T})$.
In $\mathcal{G}$:
\vspace{-2px}
\begin{itemize}
    \item $\mathcal{O} = \{o_i\}_{i=1}^{N}$ signifies $N$ objects segmented from the image $\mathcal{I}$.
    Each object is defined by $o_i = \{c, m\}$, where $c$ belongs to one of the predefined $C$ object categories, and $m$ is a binary mask in $\{0, 1\}^{H \times W}$ for this object.
    \item $\mathcal{R} = \{r_{i,j} \mid i,j \in \{1, 2, \ldots, N\}, i \neq j\}$ denotes relations with $r_{i,j}$ being the relation between $o_i$ and $o_j$. Each $r$ belongs to one of the predefined $K$ relation categories (or no relation), $N$ is the number of objects in the image. 
\end{itemize}

\begin{figure}[t]
    \centering
    \includegraphics[width=1.0\linewidth]{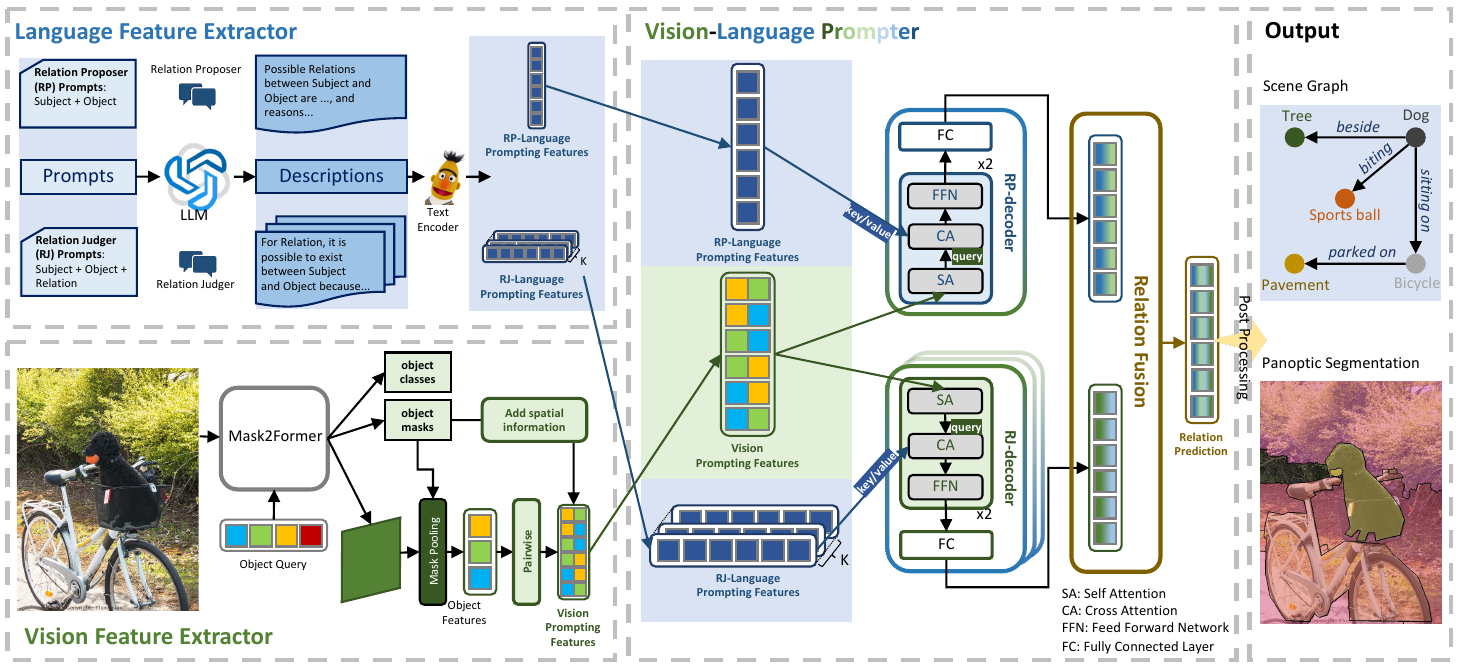}
    \vspace{-7mm}
    \caption{The overall framework of VLPrompt, which comprises three components: the vision feature extractor, the language feature extractor and the vision-language prompter.}
    \vspace{-4mm}
    \label{fig:overview}
\end{figure}

As shown in Fig.~\ref{fig:overview}, our method comprises three main components: \emph{vision feature extractor}, \emph{language feature extractor}, and \emph{vision-language prompter}.
The \emph{vision feature extractor} (Sec.~\ref{sec:vision_feature_extractor}) adapts from a segmentation network (\eg Mask2Former~\cite{cheng2022masked}) to predict object masks and form subject-object pairs for feature extraction, which result into vision prompting features.
For the \emph{language feature extractor} (Sec.~\ref{sec:language_feature_extractor}), we generate different types of descriptions by leveraging the extensive common sense knowledge embedded in LLMs through carefully designed prompts, which is beneficial for relations of different frequencies.
These descriptions are then converted into language prompting features using a text encoder. 
Next, the vision and language prompting features are fed into a \emph{vision-language prompter} (Sec.~\ref{sec:vision_language_prompter}), where vision prompting features interact with different types of language prompting features respectively, so as to take advantage of the complimentary language information to assist relation predictions. 
Finally, these relation predictions are combined by a relation fusion module 
to achieve the final relation prediction.

\subsection{Vision Feature Extractor}
\label{sec:vision_feature_extractor}

Given an image $\mathcal{I}$, we first leverage Mask2Former~\cite{cheng2022masked} to produce $N$ objects with masks.
They are formed into $N \times (N-1)$ subject-object pairs by pairing any two distinct ones. 
The purpose of the vision feature extractor is to extract vision prompting feature for each subject-object pair, which includes visual features from the segmentation network itself as well as spatial features between the subject-object pairs.
In this way, we can enhance the representations of the relations between subject-object pairs.  

\noindent \textbf{Subject-object visual features.}
We first obtain the features corresponding to each object.
Considering that the output feature map by the pixel decoder in Mask2Former retains rich information of the image, we use mask pooling to obtain object features corresponding to the $N$ objects from the feature map based on each object's mask $m$.
Then, we pair and concatenate any two distinct object features to form $N \times (N-1)$ subject-object visual features $F_{V}^{vi}$.

\noindent \textbf{Subject-object spatial features.}
To further enhance the representations of subject-object pairs, especially their spatial relations,  we are inspired by~\cite{peyre2017weakly} to encode the spatial features into the subject-object visual features.
Specifically, given $o_i$ and $o_j$ corresponding to subject and object, we first derive their encompassing bounding boxes $b_i=[x_i, y_i, w_i, h_i]$ and $b_j=[x_j, y_j, w_j, h_j]$, where $(x, y)$ is the center of the bounding box, and $(w, h)$ are the width and height.
Next we construct spatial features: 
\vspace{-5mm}
\begin{equation}
    v(o_i, o_j) = [\frac{x_{j} - x_{i}}{\sqrt{w_{i}h_{i}}}, \frac{y_{j} - y_{i}}{\sqrt{w_{i}h_{i}}}, \sqrt{\frac{w_{j}h_{j}}{w_{i}h_{i}}}, \frac{b_i \cap b_j}{b_i \cup b_j}, \frac{w_i}{h_i}, \frac{w_j}{h_j}],
\end{equation}
where $v(o_i, o_j)$ encodes the spatial relation between $o_i$ and $o_j$, such as the ratio of their bounding box sizes, the overlap between two objects and the aspect ratio of each object.
Then, we use a FC layer to expand the spatial features to the same dimension as $F_{V}^{vi}$, resulting in $F_{V}^{sp}$.

Finally, we apply a FC layer to the sum of $F_{V}^{vi}$ and $F_{V}^{sp}$ and output vision prompting features $F_V \in \mathbb{R}^{N \times (N-1) \times D_{V}}$, where $D_{V}$ is the vision feature dimension.

\vspace{-2mm}
\subsection{Language Feature Extractor}
\label{sec:language_feature_extractor}

The purpose of the language feature extractor is to leverage the extensive common sense knowledge embedded in LLMs for providing additional language information to the PSG task, which can mitigate the long-tail problem in relation prediction.
To achieve this, we need to design various prompts to elicit outputs from LLMs.
On one hand, LLMs can act as a relation proposer, suggesting possible relations between two objects, which often are frequently occurring relations in the real world.
On the other hand, LLMs can also serve as a relation judger, given a subject-object pair and their specific relation, LLMs make judgment and provide reasoning for this relation. This allows detailed descriptions even for rare relations.
Specifically, we design two types of prompts: relation proposer prompt (RP-Prompt) for proposing and explaining potential relations given a subject-object pair; relation judger prompt (RJ-Prompt) for judging and reasoning  upon a specific subject-relation-object triplet. 
Below, we detail how to obtain RP- and RJ-language prompting features based on the generated descriptions.

\noindent \textbf{RP-language prompting feature.}
For RP-language prompting feature, we stimulate LLMs to guess all possible relations between two given objects $o_i$ and $o_j$, along with explanation for these relations.
To achieve this, we utilize the chain-of-thought technique~\cite{wei2022chain}: we engage in a dialogue with an LLM (\eg GPT-3.5~\cite{chatgpt}), initially informing it to act as a relation proposer and defining the task.
Then an example is provided to the LLM to clarify its role.
We explicitly mention the predefined $K$ relations in the prompt, guiding the LLM to propose from them.
Finally, a certain subject-object pair ($o_i$ and $o_j$) is given to the relation proposer prompt.
By giving this prompt to the LLM, we obtain the description for potential relations between $o_i$ and $o_j$.
For predefined relations that are not proposed by the LLM, we would append a template phrase by the end of the description, such as ``It is not likely for $o_i$ and $o_j$ to have relation $r$.'', to make sure the language description cover all relations.
To encode the description into a feature interpretable by our model, we use a text encoder (\eg OpenAI Embeddings~\cite{chatgpt}) to convert the description into the feature $F_{L(i, j)}^{RP} \in \mathbb{R}^{1 \times D_{L}}$, namely RP-language prompting feature.
This process consolidates all descriptions into a single feature, allowing for a condensed reflection of the distinct attributes of common relations between subject and object.

\noindent \textbf{RJ-language prompting feature.}
For RJ-language prompting feature, we design the relation judger prompt: 
we not only provide two objects $o_i$, $o_j$ but also specify a relation $r_{k}$ between them. 
By using the common sense knowledge, the LLM judges whether the relation $r_{k}$ could plausibly exist between $o_i$ and $o_j$ and provides reason.
Similar to above, we use the chain-of-thought technique~\cite{wei2022chain} by telling the LLM that it serves as a relation judger; we first define the task, then give the example, and finally, the triplet ($o_i\text{-}r_{k}\text{-}o_j$) is provided.
Following the same process as for RP-language prompting feature, we feed the relation judger prompt to the LLM to obtain the language description and leverage the text encoder (same as above) to encode it into the RJ-language prompting feature $F_{L(i, j, k)}^{RJ} \in \mathbb{R}^{1 \times D_{L}}$. Different from the RP-language prompting feature, we encode each relation triplet into an individual feature, storing more detailed and subtle information for each relation, hence favouring rare relations.

By enumerating all $C$ objects and $K$ relations in the dataset, we derive all RP- and RJ-language prompting features.
To further enhance the practicality of our method and avoid repeatedly invoking LLM at runtime, we store them in a database.
Given an image with $N$ objects, we retrieve two sets of features, $F_{L}^{RP} \in \mathbb{R}^{N \times (N-1) \times D_{L}}$ and $F_{L}^{RJ} \in \mathbb{R}^{N \times (N-1) \times K \times D_{L}}$ from the database.
Note that for a specific relation $r_{k}$ between all subject object pairs in this image, the RJ-language prompting feature is denoted as $F_{L(k)}^{RJ} \in \mathbb{R}^{N \times (N-1) \times D_{L}}$.

\vspace{-2mm}
\subsection{Vision-Language Prompter}
\label{sec:vision_language_prompter}

To enable the vision prompting feature to predict relations from both macro and detailed perspectives, we let $F_V$ interact with $F_L^{RP}$ and $F_L^{RJ}$ through two separate decoders, termed as RP-decoder and RJ-decoder, responsible for the interaction from $F_V$ to $F_L^{RP}$ and $F_L^{RJ}$, respectively. 
Each decoder contains two standard transformer decoder blocks~\cite{vaswani2017attention}, followed by a FC layer for relation prediction.
The predictions from the two decoders are complementary: the RP-language prompting feature focuses on the condensed and distinct attributes of frequently occurred relations for a given subject-object pair; in contrast, the RJ-language prompting feature focuses on the detailed and subtle attributes of every possible relation (common or rare) for the subject-object pair. A relation fusion module consisting of a gating network is thereby devised by the end to fuse the relation predictions from both decoders into the final one. Before feeding $F_V$, $F_L^{RP}$, and $F_L^{RJ}$ into different decoders, we use a FC layer to transform their dimensions to a uniform dimension $D$. Below, we specify the RP-decoder, RJ-decoder, and the relation fusion module, respectively.

\noindent \textbf{RP-decoder.}
The RP-decoder aims to utilize the $F_{L}^{RP}$ to assist $F_{V}$ in relation prediction, particularly for relations that are frequently encountered between $o_i$ and $o_j$ in the real world.
In the first transformer decoder block, $F_{V}$ is firstly fed into the self-attention layer, primarily aggregating the visual relational information in $F_{V}$.
Afterwards, the self-attended $F_{V}$ as query and the $F_{L}^{RP}$ as key/value are engaged in the subsequent cross-attention layer,
 aggregating the common sense knowledge of potential relations between $o_i$ and $o_j$ into $F_{V}$. 
The output is further processed through a feed-forward network.
In the second transformer decoder block, we repeat the aforementioned process.
Finally, a fully connected layer is used to transform feature dimension $D$ to the number of relations $K$, and a sigmoid function is applied afterwards to obtain $R^{RP} \in \mathbb{R}^{N \times (N-1) \times K}$. 

\noindent \textbf{RJ-decoder.}
The RJ-decoder aims to facilitate interaction between the RJ-language prompting feature $F_{L}^{RJ}$ with the vision prompting feature $F_{V}$.
Since $F_{L}^{RJ}$ a group of individual language prompting feature for every relation triplet ($o_i\text{-}r_{k}\text{-}o_j$), $F_V$ thus has the opportunity to interact with each relation's language representation independently, which can be particularly beneficial to rare relations.
We conduct parallel interactions between $F_V$ and the $K$ triplet features contained in $F_L^{RJ}$.
For each triplet, the interaction process between $F_V$ and $F_{L(k)}^{RJ}$ is the same to that of the RP-decoder, except that the final FC layer is now only responsible for predicting the probability of certain relation between $o_i$ and $o_j$. The FC layer is used to transform the feature dimension from $D$ to 1.
Finally, we concatenate the respective outputs to get the predictions for all $K$ relations, a sigmoid function is applied over them to obtain $R^{RJ} \in \mathbb{R}^{N \times (N-1) \times K}$.

\noindent \textbf{Relation Fusion.}
Upon obtaining $R^{RP}$ and $R^{RJ}$, we aim to take the strength of both via a relation fusion module.
We devise a gating network consisting of 3-layer MLP to generate two sets of weights, $W^{RP}$ and $W^{RJ}$, each matching the shape of $R^{RP}$ and $R^{RJ}$.
We use the sum of $F_V$ and $F_{L}^{RP}$ as the input of the gating network, and output $W^{RP}$.
For $W^{RJ}$, we use the sum of $F_V$ and the mean of $F_{L}^{RJ}$ along the relation dimension as input to the gating network.
$W^{RP}$ and $W^{RJ}$ are used to element-wisely multiply with $R^{RP}$ and $R^{RJ}$ respectively and the final relation prediction $R$ is a weighted combination:
\vspace{-1mm}
\begin{equation}
    R = W^{RP} \odot R^{RP} + W^{RJ} \odot R^{RJ},
\end{equation}
where $\odot$ is element-wise multiplication, and $R \in \mathbb{R}^{N \times (N-1) \times K}$.

Finally, the prediction $R$, combined with the object categories and masks predicted by the vision feature extractor, forms the panoptic scene graph $\mathcal{G}$.

\vspace{-2mm}
\subsection{Model Training}
\label{sec:model_training}
\vspace{-2mm}
Our model training comprises two parts.
The first part is the segmentation loss $\mathcal{L}_{seg}$ used in the vision feature extractor for panoptic segmentation, we simply follow the loss used in~\cite{cheng2022masked}.
The second part is the relation loss. Since the same subject-object pair might have multiple relations, we use a binary cross-entropy loss~\cite{su2022zlpr}.
To effectively train the vision-language prompter, we apply the relation loss separately to $R^{RP}$, $R^{RJ}$, and $R$ and sum them up as the final relation loss, denoted by  $\mathcal{L}_{rel}$.
The final loss $\mathcal{L}$ is
\vspace{-1mm}
\begin{equation}
    \mathcal{L} = \lambda \mathcal{L}_{seg} + \mathcal{L}_{rel} \; ,
\end{equation}
where $\lambda$ is the weighting coefficient.
In the language feature extractor, we directly utilize pre-trained LLMs, thus eliminating the need for additional model training.

\vspace{-2mm}
\section{Experiments}
\label{sec:experiments}
\vspace{-2mm}

\vspace{-2px}
\subsection{Datasets}
\vspace{-2px}

\textbf{Panoptic Scene Graph (PSG) dataset}~\cite{yang2022panoptic}.
This is the first dataset dedicated to the PSG task, comprising 48,749 annotated images, including 2,186 test images and 46,563 training images.
The dataset includes 80 ``thing''~\cite{lin2014microsoft} and 53 ``stuff'' categories~\cite{caesar2018coco}, as well as 56 relation categories.

\noindent \textbf{Visual Genome (VG) dataset}~\cite{krishna2017visual}.
This dataset is widely used in the SGG task.
To validate our method, we also conduct experiments on the VG dataset.
Following previous works~\cite{zellers2018neural, chen2019counterfactual}, we use the VG-150 variant, which includes 150 object categories and 50 relation categories.

For details on the sub-tasks, evaluation metrics, and specific implementation details for these two datasets, please refer to the supplementary material.

\vspace{-4px}
\subsection{Comparison to the state-of-the-art}
\noindent \textbf{PSG.}
Tab.~\ref{tab:psg_results} reports the performance of our method compared to previous state-of-the-art methods on the PSG dataset~\cite{yang2022panoptic}.
Previous methods rely solely on vision inputs, \ie images, while ours utilizes both vision and language inputs.
For a fair comparison, we use the same Resnet-50~\cite{he2016deep} backbone in for all methods in the vision feature extractor.
Our method shows superior performance compared to all previous methods.
Particularly, it outperforms the previous best-performing method \cite{zhou2023hilo} by a large margin, \eg +9.4\% in R@100 and +20.6\% in mR@100.
It is noteworthy that we achieve substantial improvements in the mR@K metric, demonstrating the significant benefits of adding language information for predicting rare relations.

\begin{table}[t]\small
    \centering
    \vspace{-4mm}
    \caption{Comparison between our VLPrompt and other methods on the PSG dataset.
    Our method shows superior performance compared to all previous methods.}
    \vspace{-2mm}
    \begin{tabular}{l|c|cc|cc|cc}
    \hline
        ~ & ~ & \multicolumn{6}{c}{Scene Graph Detection} \\
        \cline{3-8}
        Method & Model Input & R@20 & mR@20 & R@50 & mR@50 & R@100 & mR@100 \\
        \hline
        IMP~\cite{xu2017scene} & Vision & 16.5 & 6.5 & 18.2 & 7.1 & 18.6 & 7.2 \\
        MOTIF~\cite{zellers2018neural} & Vision & 20.0 & 9.1 & 21.7 & 9.6 & 22.0 & 9.7 \\
        VCTree~\cite{tang2019learning} & Vision & 20.6 & 9.7 & 22.1 & 10.2 & 22.5 & 10.2 \\
        GPSNet~\cite{lin2020gps} & Vision & 17.8 & 7.0 & 19.6 & 7.5 & 20.1 & 7.7 \\
        PSGTR~\cite{yang2022panoptic} & Vision & 28.4 & 16.6 & 34.4 & 20.8 & 36.3 & 22.1 \\
        PSGFormer~\cite{yang2022panoptic} & Vision & 18.0 & 14.8 & 19.6 & 17.0 & 20.1 & 17.6 \\
        PairNet~\cite{wang2023pair} & Vision & 29.6 & 24.7 & 35.6 & 28.5 & 39.6 & 30.6 \\
        HiLo~\cite{zhou2023hilo} & Vision & 34.1 & 23.7 & 40.7 & 30.3 & 43.0 & 33.1 \\
        \hline
        \textbf{VLPrompt (ours)} & Vision + Language & \textbf{39.4} & \textbf{34.7} & \textbf{47.6} & \textbf{45.1} & \textbf{52.4} & \textbf{53.7} \\
        \hline
    \end{tabular}
    \vspace{-8mm}
    \label{tab:psg_results}
\end{table}

\begin{minipage}{\linewidth}
\begin{minipage}[t]{0.535\linewidth}
\centering
    \makeatletter\def\@captype{table}\makeatother\caption{Alleviation the long-tail problem.}
    \label{tab:long_tail}
    \resizebox{1.0\linewidth}{!}{
    \begin{tabular}{l|cc|cc}
    \hline
        ~ & \multicolumn{2}{c|}{Common relations} & \multicolumn{2}{c}{Rare relations}\\
        ~ & w/o LLM & w/ LLM & w/o LLM & w/ LLM\\ \hline
        mR@100 & 54.7 & 57.0 (\textcolor{darkgreen}{+2.3}) & 37.9 & 51.7 (\textcolor{darkgreen}{+13.8}) \\ \hline
    \end{tabular}
    }
\end{minipage}
\hfill
\begin{minipage}[t]{0.46\linewidth}
\centering
    \makeatletter\def\@captype{table}\makeatother\caption{Comparison with GPT-4V in PSG.}
    \label{tab:gpt_4v}
    \resizebox{0.98\linewidth}{!}{
    \begin{tabular}{l|ccc}
    \hline
        Method & R/mR@20 & R/mR@50 & R/mR@100 \\
        \hline
        VLPrompt & 39.4/34.7 & 47.6/45.1 & 52.4/53.7 \\
        GPT-4V & 23.6/15.6 & 27.5/16.8 & 35.2/22.5 \\
        \hline
    \end{tabular}
    }
\end{minipage}
\end{minipage}



\noindent \textit{For the long-tail problem.}
In PSG task, mean recall is often used as a reflection for a method's ability to solve long-tail problem~\cite{tang2020unbiased}.
To further validate it, in PSG dataset, we split relations occurring over 1000 times as common relations, and those under 1000 as rare relations, resulting into 21 common and 35 rare relations.
As shown in Tab.~\ref{tab:long_tail}, our method's integration of language information leads to 2.3 increase in mR@100 for common relations and 13.8 boost for rare relations.
The significant improvement in the latter highlights the effectiveness in enhancing rare relation prediction.
Note, w/o LLM is our method without language information (Sec.~\ref{sec:method}).

\noindent \textit{Comparison with large multimodal models.}
Large multimodal models (LMMs) currently demonstrate great performance across various multimodal tasks and can also predict relations with specific instructions.
To further validate the performance of LMMs on PSG task, we test a state-of-the-art LMMs GPT-4V on the PSG test set.
To ensure fairness in comparison, we allow GPT-4V to use the segmentation results output by our method, thus ensuring same segmentation performance.
Specifically, following the approach of \cite{yang2023setofmark}, we attach the panoptic segmentation results extracted by our model to the original image and input it to GPT-4V.
This allows GPT-4V to obtain object segmentation results and the corresponding categories consistent with our method.
Subsequently, we use GPT-4V to predict the relations for all object pairs.
The results (GPT-4V), as shown in Tab.~\ref{tab:gpt_4v}, indicate that the general LMMs are not dedicated to the PSG task thus can not work very well.

\noindent \textbf{VG.}
Tab.~\ref{tab:vg_results} reports the performance of our method compared to previous methods on the VG dataset~\cite{krishna2017visual}.
To adapt our method to the VG dataset, \ie a bounding box-based SGG task, we first use a Segment Anything Model (SAM)~\cite{kirillov2023segment} with VG dataset's ground-truth bounding box annotations as prompts to transform the VG dataset into a dataset suitable for instance segmentation tasks.
We then train a Mask2Former on this instance segmentation dataset, enabling our method to be adapted to the VG dataset.
As shown in Tab.~\ref{tab:vg_results}, our method surpasses previous vision-only and vision-language models in mR@K, indicating that incorporating language information effectively enhances the prediction performance for rare relations and alleviates the long-tail problem.

\begin{figure}[t]
    \centering
    \includegraphics[width=1.0\linewidth]{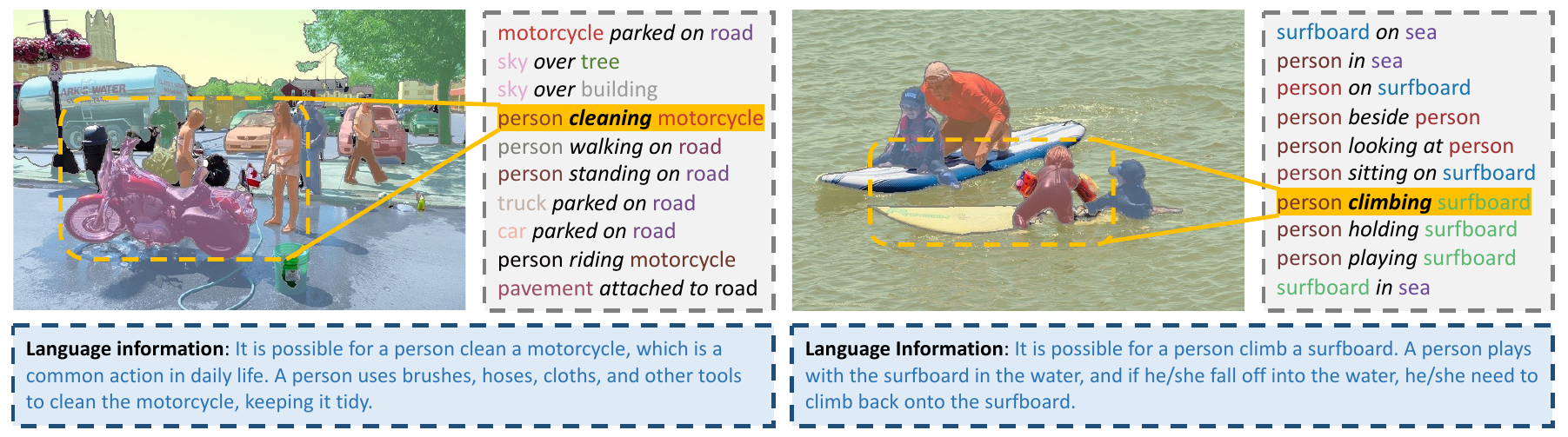}
    \vspace{-6mm}
    \caption{Visualization results of our VLPrompt.
    We show two examples.
    For each example, the top left displays the predicted segmentation results, the top right shows the top 10 predicted relation triplets (all are correct relation triplets), and bottom is the language snippet utilized for predicting the highlighted triplets in yellow.
    }
    \label{fig:vis}
    \vspace{-2mm}
\end{figure}

\noindent \textbf{Qualitative analysis.}
As shown in Fig.~\ref{fig:vis}, with the inclusion of language information, VLPrompt successfully predicts challenging relations, which are the highlighted in yellow.



\vspace{-2mm}
\subsection{Ablation study}
\vspace{-2mm}

\subsubsection{Vision Feature Extractor}

\noindent \textbf{Object features from pixel decoder.}
To validate the superiority of obtaining object features from the pixel decoder (PixelDec) of Mask2Former, we experiment with an alternative approach: acquiring corresponding object features from the transformer decoder (TsfmDec) of Mask2Former. This is a common practice in the literature~\cite{cong2023reltr}.
Experiments (PixelDec$\rightarrow$TsfmDec) in Tab.~\ref{tab:ablation_vision_feature_extractor} show that the feature from the pixel decoder performs 3.4\% better in mR@100 than that from the transformer decoder, as the pixel decoder contains more comprehensive vision information.

\begin{table}[h]\small
    \centering
    \caption{Comparison between our VLPrompt and other methods on the VG dataset.}
    \vspace{-2mm}
    \resizebox{1.0\linewidth}{!}{
    \begin{tabular}{l|c|cc|cc|cc}
    \hline
        ~ & ~ & \multicolumn{6}{c}{Scene Graph Detection} \\
        \cline{3-8}
        Method & Model Input & R@20 & mR@20 & R@50 & mR@50 & R@100 & mR@100 \\
        \hline
        MOTIF~\cite{zellers2018neural} & Vision & 21.7 & ~--~ & 31.0 & 6.7 & 35.1 & 7.7 \\
        VCTree~\cite{tang2019learning} & Vision & 22.0 & ~--~ & 30.2 & 6.7 & 34.6 & 8.0 \\
        Transformer~\cite{tang2020unbiased} & Vision & ~--~ & ~--~ & 30.0 & 7.4 & 34.3 & 8.8 \\
        GPSNet~\cite{lin2020gps} & Vision & 22.3 & ~--~ & 30.3 & 5.9 & 35.0 & 7.1 \\
        GPSNet + IETrans~\cite{zhang2022fine} & Vision & ~--~ & ~--~ & 25.9 & 14.6 & 28.1 & 16.5 \\
        GPSNet + HiLo~\cite{zhou2023hilo} & Vision & ~--~ & ~--~ & 25.6 & 15.8 & 27.9 & 18.0 \\
        \hline
        SVRP~\cite{he2022towards} & Vision + Language & ~--~ & ~--~ & \textbf{31.8} & 10.5 & \textbf{35.8} & 12.8 \\
        \textbf{VLPrompt (ours)} & Vision + Language & \textbf{24.0} & \textbf{10.7} & 28.7 & \textbf{17.5} & 31.6 & \textbf{20.2} \\
        \hline
    \end{tabular}
    }
    \vspace{-8mm}
    \label{tab:vg_results}
\end{table}


\noindent \textbf{Mask pooling for object feature extraction.}
Common methods to obtain object features from the pixel decoder include mask pooling (MaskPool) and bounding box pooling (BboxPool)~\cite{girshick2015fast}.
We validate our choice of mask pooling through ablation experiments (MaskPool$\rightarrow$BboxPool).
As shown in Tab.~\ref{tab:ablation_vision_feature_extractor}, the performance using mask pooling is 1.1\% higher in mR@100 than that using bbox pooling.
Mask pooling is more suitable for the mask prediction in the PSG task.

\begin{minipage}{\linewidth}
\begin{minipage}[t]{0.51\linewidth}
\centering
\makeatletter\def\@captype{table}\makeatother\caption{Ablation study for the Vision Feature Extractor.}
    \label{tab:ablation_vision_feature_extractor}
    \resizebox{1.0\linewidth}{!}{
    \begin{tabular}{l|ccc}
    \hline
        Method & R/mR@20 & R/mR@50 & R/mR@100 \\
        \hline
        VLPrompt & 39.4 / 34.7 & 47.6 / 45.1 & 52.4 / 53.7 \\
        \hdashline
        PixelDec $\rightarrow$ TsfmDec & 37.4 / 32.1 & 45.8 / 43.7 & 50.1 / 50.3 \\
        MaskPool $\rightarrow$ BboxPool & 38.6 / 33.7 & 46.4 / 44.8 & 51.0 / 52.6 \\
        Concat. $\rightarrow$ Sub. & 35.4 / 29.8 & 44.1 / 42.3 & 48.6 / 48.9 \\
        w/o Spatial Feat. & 39.0 / 33.0 & 46.6 / 44.3 & 51.7 / 52.1 \\
        \hline
    \end{tabular}
    }
\end{minipage}
\hfill
\begin{minipage}[t]{0.47\linewidth}
\centering
\makeatletter\def\@captype{table}\makeatother\caption{Ablation study for the Language Feature Extractor.}
    \label{tab:ablation_language_feature_extractor}
    \resizebox{1.0\linewidth}{!}{
    \begin{tabular}{l|ccc}
    \hline
        Method & R/mR@20 & R/mR@50 & R/mR@100 \\
        \hline
        VLPrompt & 39.4/34.7 & 47.6/45.1 & 52.4/53.7 \\
        \hdashline
        w/o CoT & 36.8/27.9 & 44.3/38.8 & 48.6/45.7 \\
        Ext$^\text{RP} \rightarrow$ Ext$^\text{RJ}$ & 39.1/34.2 & 47.5/45.1 & 52.3/53.5 \\
        Ext$^\text{RJ} \rightarrow$ Ext$^\text{RP}$ & 38.9/31.2 & 46.8/41.7 & 51.0/49.9 \\
        Swap(Ext$^\text{RP}$, Ext$^\text{RJ}$) & 36.2/29.4 & 44.3/38.6 & 48.7/46.4 \\
        \hline
    \end{tabular}
    }
\end{minipage}
\end{minipage}


\noindent \textbf{Concatenating object features into a pair feature.}
Common methods for merging the features of two objects for their relation prediction include concatenation (Concat.) \cite{zhang2019graphical} and subtraction (Sub.) \cite{cheng2022visual}.
Notably, addition of the features cannot be used due to the inherent order requested between the subject and object.
Results (Concat.$\rightarrow$Sub.) in Tab.~\ref{tab:ablation_vision_feature_extractor} indicate that concatenation outperforms subtraction by 4.8\% on the mR@100.

\noindent \textbf{Spatial feature.}
To validate the effect of adding spatial features, we conduct an ablation study by removing these features.
The results (w/o Spatial Feat.) in Tab.~\ref{tab:ablation_vision_feature_extractor} show that this leads to a decrease of 0.7\% in R@100 and 1.6\% in mR@100.
This is because spatial features provide the model with additional spatial interaction between the subject and object (Sec.~\ref{sec:vision_feature_extractor}), thereby enhancing the vision feature for relation prediction.

\vspace{-2mm}
\subsubsection{Language Feature Extractor}

\noindent \textbf{Chain-of-thought for prompt.}
By carefully designing prompts with the chain-of-thought technique, LLMs can produce rich and accurate descriptions.
However, if we do not use chain-of-thought and instead directly ask LLMs questions, such as replacing the relation proposer prompt with ``What are the possible relations between subject and object? And why?'' or the relation judger prompt with ``Could this relation be possible between subject and object? Why?", we find that the outputs from LLMs become much less predictable and often not as expected.
We experiment without chain-of-thought (w/o CoT) and the results in Tab.~\ref{tab:ablation_language_feature_extractor} show that the performance of relation prediction significantly drops, \ie a 8.0\% decrease in mR@100.
This further illustrates the rationale and importance of using chain-of-thought technology for designing prompts.

\noindent \textbf{Different feature extraction methods.}
To validate our design of different feature extract methods for RP- and RJ-language prompting features, we study their effects.
We offer three variants:
1) For RP-language prompting features, we adopt the same way as the RJ-language prompting feature to extract feature for each relation triplet individually, we denote this variant as Ext$^\text{RP} \rightarrow$ Ext$^\text{RJ}$ in the

\begin{minipage}{\linewidth}
\begin{minipage}[t]{0.48\linewidth}
\centering
    \makeatletter\def\@captype{table}\makeatother\caption{Ablation study for the Vision-Language Prompter.}
    \label{tab:ablation_vision_language_prompter}
    \resizebox{0.99\linewidth}{!}{
    \begin{tabular}{l|ccc}
    \hline
        Method & R/mR@20 & R/mR@50 & R/mR@100 \\
        \hline
        VLPrompt & 39.4/34.7 & 47.6/45.1 & 52.4/53.7 \\
        \hdashline
        w/o Language & 37.1/26.8 & 45.9/37.2 & 50.0/44.2 \\
        RP-decoder & 38.4/30.6 & 46.6/41.2 & 50.9/49.6 \\
        RJ-decoder & 37.9/34.2 & 45.3/44.6 & 50.4/52.8 \\
        \hline
    \end{tabular}
    }
\end{minipage}
\hfill
\begin{minipage}[t]{0.50\linewidth}
\centering
    \makeatletter\def\@captype{table}\makeatother\caption{Ablation study for the number of decoder blocks.}
    \label{tab:ablation_decoder_block_number}
    \vspace{+2px}
    \resizebox{1.0\linewidth}{!}{
    \begin{tabular}{c|ccc}
    \hline
        Block Number & R/mR@20 & R/mR@50 & R@/mR@100 \\
        \hline
        1 & 35.1/29.8 & 44.8/39.5 & 47.2/46.9 \\
        2 & \textbf{39.4}/34.7 & \textbf{47.6}/45.1 & \textbf{52.4}/53.7 \\
        4 & 38.9/\textbf{34.8} & 47.4/\textbf{45.2} & 52.1/\textbf{53.9} \\
        12 & 18.4/14.5 & 27.5/23.0 & 33.2/29.4 \\
        \hline
    \end{tabular}
    }
\end{minipage}
\end{minipage}



\noindent Tab.~\ref{tab:ablation_language_feature_extractor}.
It shows that performance is slightly lower than our original VLPrompt, but with increased computation.
2) For RJ-language prompting features, we adopt the same way as the RP-language prompting feature to extract all relation triplets into one feature, denoted by Ext$^\text{RJ} \rightarrow$ Ext$^\text{RP}$ in the Tab.~\ref{tab:ablation_language_feature_extractor}.
We observe a clear drop in mR@K, as this would condense the features of relations between subject and object while dropping the subtle details, which can be especially disadvantageous for rare relations.
3) We swap the feature extraction methods between Ext$^\text{RP}$ and Ext$^\text{RJ}$, denoted by Swap(Ext$^\text{RP}$, Ext$^\text{RJ}$) in Tab.~\ref{tab:ablation_language_feature_extractor}.
We observe a substantial decrease in both metrics.

Our original feature extraction is specifically designed on one side to focus on the condensed and distinct attributes of commonly occurring relations; on the other side to focus on the detailed and subtle attributes of all possible especially rare relations for a subject-object pair.

\subsubsection{Vision-Language Prompter}

\noindent \textbf{Language information.}
To validate the efficacy of incorporating language information, we conduct a comparative experiment by replacing all language prompting features in the vision-language prompter with vision prompting features.
This approach ensures that all other factors remain constant while assessing the impact of language information.
As shown in Tab.~\ref{tab:ablation_vision_language_prompter}, the experimental results (w/o Language) reveal a significant decrease in the mR@100 by 9.5\% when language information is removed, which demonstrates the substantial impact of language information in mitigating the long-tail problem.

\noindent \textbf{Results of RP-decoder and RJ-decoder.}
We test the relation prediction performance of both RP-decoder and RJ-decoder separately.
The results (RP-decoder and RJ-decoder) in Tab.~\ref{tab:ablation_vision_language_prompter} show that the RP-decoder outperforms the RJ-decoder in the R@K (\eg 50.9\% \vs 50.4\% for R@100), while the RP-decoder scores lower in mR@K compared to the RJ-decoder (\eg 49.6\% \vs 52.8\% for mR@100).
This indicates that the RP-decoder is good at predicting frequent relation classes, whereas the RJ-decoder excels in rare relation classes, which indicates that they are complementary.


\noindent \textbf{The number of decoder blocks.}
To elucidate the reason that both RP-decoder and RJ-decoder use a 2-block transformer decoder, we vary different numbers of blocks in Tab.~\ref{tab:ablation_decoder_block_number}.
We observe that when increasing the transformer decoder blocks from 1 to 2, there is a significant improvement in model performance.
However, increasing from 2 to 4 blocks does not change much for the performance.
Further increasing the decoder blocks to 12 leads to a notable decrease in performance, suggesting that too many decoder blocks may cause the model to overfit.
Considering both performance and speed, we choose 2 blocks.

\section{Conclusion and Limitation}
\label{sec:conclusion}

In this work, we introduce VLPrompt, the first method to incorporate language information generated by LLMs to enhance the PSG task performance.
VLPrompt utilizes the chain-of-thought method in designing prompts, enabling LLMs to generate rich descriptions for relation prediction. 
Additionally, we develop a prompter network based on attention mechanisms to facilitate comprehensive interaction between vision and language information, achieving high-quality relation prediction.
Experiments demonstrate that our method significantly outperforms the current state-of-the-art on the PSG dataset and mitigates the long-tail problem for relations.
In future work, we plan to explore the use of LLMs for open-set relation prediction and further refine the model by distillation to enhance efficiency, enabling broader application in downstream tasks.

\noindent \textit{Limitation.} Although our method achieves significant performance improvements on the PSG dataset, it requires pre-extraction of language information from LLMs for unbiased relation prediction. 



{\small
\bibliographystyle{plainnat}
\bibliography{neurips_2024}

\begin{thebibliography}{78}
\providecommand{\natexlab}[1]{#1}
\providecommand{\url}[1]{\texttt{#1}}
\expandafter\ifx\csname urlstyle\endcsname\relax
  \providecommand{\doi}[1]{doi: #1}\else
  \providecommand{\doi}{doi: \begingroup \urlstyle{rm}\Url}\fi

\bibitem[Aditya et~al.(2018)Aditya, Yang, Baral, Aloimonos, and Ferm{\"u}ller]{aditya2018image}
Somak Aditya, Yezhou Yang, Chitta Baral, Yiannis Aloimonos, and Cornelia Ferm{\"u}ller.
\newblock Image understanding using vision and reasoning through scene description graph.
\newblock \emph{Computer Vision and Image Understanding}, pages 33--45, 2018.

\bibitem[Amiri et~al.(2022)Amiri, Chandan, and Zhang]{amiri2022reasoning}
Saeid Amiri, Kishan Chandan, and Shiqi Zhang.
\newblock Reasoning with scene graphs for robot planning under partial observability.
\newblock \emph{IEEE Robotics and Automation Letters}, pages 5560--5567, 2022.

\bibitem[Anthropic(2023)]{claude}
Anthropic.
\newblock Claude, 2023.
\newblock URL \url{https://claude.ai/chats}.

\bibitem[Caesar et~al.(2018)Caesar, Uijlings, and Ferrari]{caesar2018coco}
Holger Caesar, Jasper Uijlings, and Vittorio Ferrari.
\newblock Coco-stuff: Thing and stuff classes in context.
\newblock In \emph{CVPR}, 2018.

\bibitem[Carion et~al.(2020)Carion, Massa, Synnaeve, Usunier, Kirillov, and Zagoruyko]{carion2020end}
Nicolas Carion, Francisco Massa, Gabriel Synnaeve, Nicolas Usunier, Alexander Kirillov, and Sergey Zagoruyko.
\newblock End-to-end object detection with transformers.
\newblock In \emph{ECCV}, 2020.

\bibitem[Chen et~al.(2023)Chen, Li, Shen, Yang, Li, Keutzer, Darrell, and Liu]{chen2023large}
Liangyu Chen, Bo~Li, Sheng Shen, Jingkang Yang, Chunyuan Li, Kurt Keutzer, Trevor Darrell, and Ziwei Liu.
\newblock Large language models are visual reasoning coordinators.
\newblock \emph{arXiv preprint arXiv:2310.15166}, 2023.

\bibitem[Chen et~al.(2019)Chen, Zhang, Xiao, He, Pu, and Chang]{chen2019counterfactual}
Long Chen, Hanwang Zhang, Jun Xiao, Xiangnan He, Shiliang Pu, and Shih-Fu Chang.
\newblock Counterfactual critic multi-agent training for scene graph generation.
\newblock In \emph{ICCV}, 2019.

\bibitem[Chen et~al.(2020)Chen, Jin, Wang, and Wu]{chen2020say}
Shizhe Chen, Qin Jin, Peng Wang, and Qi~Wu.
\newblock Say as you wish: Fine-grained control of image caption generation with abstract scene graphs.
\newblock In \emph{CVPR}, 2020.

\bibitem[Cheng et~al.(2022{\natexlab{a}})Cheng, Misra, Schwing, Kirillov, and Girdhar]{cheng2022masked}
Bowen Cheng, Ishan Misra, Alexander~G Schwing, Alexander Kirillov, and Rohit Girdhar.
\newblock Masked-attention mask transformer for universal image segmentation.
\newblock In \emph{CVPR}, 2022{\natexlab{a}}.

\bibitem[Cheng et~al.(2022{\natexlab{b}})Cheng, Wang, Wu, Hu, Jeon, Tao, and Zhou]{cheng2022visual}
Jun Cheng, Lei Wang, Jiaji Wu, Xiping Hu, Gwanggil Jeon, Dacheng Tao, and Mengchu Zhou.
\newblock Visual relationship detection: A survey.
\newblock \emph{IEEE Transactions on Cybernetics}, pages 8453--8466, 2022{\natexlab{b}}.

\bibitem[Cong et~al.(2023)Cong, Yang, and Rosenhahn]{cong2023reltr}
Yuren Cong, Michael~Ying Yang, and Bodo Rosenhahn.
\newblock Reltr: Relation transformer for scene graph generation.
\newblock \emph{IEEE Transactions on Pattern Analysis and Machine Intelligence}, 2023.

\bibitem[Dai et~al.(2017)Dai, Zhang, and Lin]{dai2017detecting}
Bo~Dai, Yuqi Zhang, and Dahua Lin.
\newblock Detecting visual relationships with deep relational networks.
\newblock In \emph{CVPR}, 2017.

\bibitem[Deng et~al.(2022)Deng, Li, Zhang, Xiang, Wang, Chen, and Ma]{deng2022hierarchical}
Youming Deng, Yansheng Li, Yongjun Zhang, Xiang Xiang, Jian Wang, Jingdong Chen, and Jiayi Ma.
\newblock Hierarchical memory learning for fine-grained scene graph generation.
\newblock In \emph{ECCV}, 2022.

\bibitem[Devlin et~al.(2018)Devlin, Chang, Lee, and Toutanova]{devlin2018bert}
Jacob Devlin, Ming-Wei Chang, Kenton Lee, and Kristina Toutanova.
\newblock Bert: Pre-training of deep bidirectional transformers for language understanding.
\newblock \emph{arXiv preprint arXiv:1810.04805}, 2018.

\bibitem[Dong et~al.(2022)Dong, Gan, Song, Wu, Cheng, and Nie]{dong2022stacked}
Xingning Dong, Tian Gan, Xuemeng Song, Jianlong Wu, Yuan Cheng, and Liqiang Nie.
\newblock Stacked hybrid-attention and group collaborative learning for unbiased scene graph generation.
\newblock In \emph{CVPR}, 2022.

\bibitem[Dupty et~al.(2020)Dupty, Zhang, and Lee]{dupty2020visual}
Mohammed~Haroon Dupty, Zhen Zhang, and Wee~Sun Lee.
\newblock Visual relationship detection with low rank non-negative tensor decomposition.
\newblock In \emph{AAAI}, 2020.

\bibitem[Gao et~al.(2018)Gao, Wang, and Wang]{gao2018image}
Lizhao Gao, Bo~Wang, and Wenmin Wang.
\newblock Image captioning with scene-graph based semantic concepts.
\newblock In \emph{ICMLC}, 2018.

\bibitem[Girshick(2015)]{girshick2015fast}
Ross Girshick.
\newblock Fast r-cnn.
\newblock In \emph{ICCV}, 2015.

\bibitem[Google(2023)]{bard}
Google.
\newblock Bard, 2023.
\newblock URL \url{https://bard.google.com/chat}.

\bibitem[Gu et~al.(2019)Gu, Zhao, Lin, Li, Cai, and Ling]{gu2019scene}
Jiuxiang Gu, Handong Zhao, Zhe Lin, Sheng Li, Jianfei Cai, and Mingyang Ling.
\newblock Scene graph generation with external knowledge and image reconstruction.
\newblock In \emph{CVPR}, 2019.

\bibitem[He et~al.(2016)He, Zhang, Ren, and Sun]{he2016deep}
Kaiming He, Xiangyu Zhang, Shaoqing Ren, and Jian Sun.
\newblock Deep residual learning for image recognition.
\newblock In \emph{CVPR}, 2016.

\bibitem[He et~al.(2022)He, Gao, Song, and Li]{he2022towards}
Tao He, Lianli Gao, Jingkuan Song, and Yuan-Fang Li.
\newblock Towards open-vocabulary scene graph generation with prompt-based finetuning.
\newblock In \emph{ECCV}, 2022.

\bibitem[Hildebrandt et~al.(2020)Hildebrandt, Li, Koner, Tresp, and G{\"u}nnemann]{hildebrandt2020scene}
Marcel Hildebrandt, Hang Li, Rajat Koner, Volker Tresp, and Stephan G{\"u}nnemann.
\newblock Scene graph reasoning for visual question answering.
\newblock \emph{arXiv preprint arXiv:2007.01072}, 2020.

\bibitem[Hu et~al.(2019)Hu, Chen, Chen, Zhang, and Gu]{hu2022neural}
Yue Hu, Siheng Chen, Xu~Chen, Ya~Zhang, and Xiao Gu.
\newblock Neural message passing for visual relationship detection.
\newblock In \emph{ICMLW}, 2019.

\bibitem[Huang et~al.(2023)Huang, Huang, Zhang, Tian, Feng, Zhang, Xie, Li, and Zhang]{huang2023inject}
Xinyu Huang, Yi-Jie Huang, Youcai Zhang, Weiwei Tian, Rui Feng, Yuejie Zhang, Yanchun Xie, Yaqian Li, and Lei Zhang.
\newblock Inject semantic concepts into image tagging for open-set recognition.
\newblock \emph{arXiv preprint arXiv:2310.15200}, 2023.

\bibitem[Hung et~al.(2020)Hung, Mallya, and Lazebnik]{hung2020contextual}
Zih-Siou Hung, Arun Mallya, and Svetlana Lazebnik.
\newblock Contextual translation embedding for visual relationship detection and scene graph generation.
\newblock \emph{IEEE transactions on pattern analysis and machine intelligence}, pages 3820--3832, 2020.

\bibitem[Hwang et~al.(2018)Hwang, Ravi, Tao, Kim, Collins, and Singh]{hwang2018tensorize}
Seong~Jae Hwang, Sathya~N Ravi, Zirui Tao, Hyunwoo~J Kim, Maxwell~D Collins, and Vikas Singh.
\newblock Tensorize, factorize and regularize: Robust visual relationship learning.
\newblock In \emph{CVPR}, 2018.

\bibitem[Kang and Yoo(2023)]{kang2023skew}
Haeyong Kang and Chang~D Yoo.
\newblock Skew class-balanced re-weighting for unbiased scene graph generation.
\newblock \emph{Machine Learning and Knowledge Extraction}, pages 287--303, 2023.

\bibitem[Kirillov et~al.(2019)Kirillov, He, Girshick, Rother, and Doll{\'a}r]{kirillov2019panoptic}
Alexander Kirillov, Kaiming He, Ross Girshick, Carsten Rother, and Piotr Doll{\'a}r.
\newblock Panoptic segmentation.
\newblock In \emph{CVPR}, 2019.

\bibitem[Kirillov et~al.(2023)Kirillov, Mintun, Ravi, Mao, Rolland, Gustafson, Xiao, Whitehead, Berg, Lo, et~al.]{kirillov2023segment}
Alexander Kirillov, Eric Mintun, Nikhila Ravi, Hanzi Mao, Chloe Rolland, Laura Gustafson, Tete Xiao, Spencer Whitehead, Alexander~C Berg, Wan-Yen Lo, et~al.
\newblock Segment anything.
\newblock \emph{arXiv preprint arXiv:2304.02643}, 2023.

\bibitem[Krishna et~al.(2017)Krishna, Zhu, Groth, Johnson, Hata, Kravitz, Chen, Kalantidis, Li, Shamma, et~al.]{krishna2017visual}
Ranjay Krishna, Yuke Zhu, Oliver Groth, Justin Johnson, Kenji Hata, Joshua Kravitz, Stephanie Chen, Yannis Kalantidis, Li-Jia Li, David~A Shamma, et~al.
\newblock Visual genome: Connecting language and vision using crowdsourced dense image annotations.
\newblock \emph{International journal of computer vision}, pages 32--73, 2017.

\bibitem[Lai et~al.(2023)Lai, Tian, Chen, Li, Yuan, Liu, and Jia]{lai2023lisa}
Xin Lai, Zhuotao Tian, Yukang Chen, Yanwei Li, Yuhui Yuan, Shu Liu, and Jiaya Jia.
\newblock Lisa: Reasoning segmentation via large language model.
\newblock \emph{arXiv preprint arXiv:2308.00692}, 2023.

\bibitem[Li et~al.(2023)Li, Ji, Wu, Li, Qin, Wei, and Zimmermann]{li2023panoptic}
Li~Li, Wei Ji, Yiming Wu, Mengze Li, You Qin, Lina Wei, and Roger Zimmermann.
\newblock Panoptic scene graph generation with semantics-prototype learning.
\newblock \emph{arXiv preprint arXiv:2307.15567}, 2023.

\bibitem[Li et~al.(2021)Li, Zhang, Wan, and He]{li2021bipartite}
Rongjie Li, Songyang Zhang, Bo~Wan, and Xuming He.
\newblock Bipartite graph network with adaptive message passing for unbiased scene graph generation.
\newblock In \emph{CVPR}, 2021.

\bibitem[Li et~al.(2022)Li, Zhang, and He]{li2022sgtr}
Rongjie Li, Songyang Zhang, and Xuming He.
\newblock Sgtr: End-to-end scene graph generation with transformer.
\newblock In \emph{CVPR}, 2022.

\bibitem[Li et~al.(2017{\natexlab{a}})Li, Ouyang, Wang, and Tang]{li2017vip}
Yikang Li, Wanli Ouyang, Xiaogang Wang, and Xiao'ou Tang.
\newblock Vip-cnn: Visual phrase guided convolutional neural network.
\newblock In \emph{CVPR}, 2017{\natexlab{a}}.

\bibitem[Li et~al.(2017{\natexlab{b}})Li, Ouyang, Zhou, Wang, and Wang]{li2017scene}
Yikang Li, Wanli Ouyang, Bolei Zhou, Kun Wang, and Xiaogang Wang.
\newblock Scene graph generation from objects, phrases and region captions.
\newblock In \emph{ICCV}, 2017{\natexlab{b}}.

\bibitem[Liao et~al.(2019)Liao, Rosenhahn, Shuai, and Ying~Yang]{liao2019natural}
Wentong Liao, Bodo Rosenhahn, Ling Shuai, and Michael Ying~Yang.
\newblock Natural language guided visual relationship detection.
\newblock In \emph{CVPRW}, 2019.

\bibitem[Lin et~al.(2014)Lin, Maire, Belongie, Hays, Perona, Ramanan, Doll{\'a}r, and Zitnick]{lin2014microsoft}
Tsung-Yi Lin, Michael Maire, Serge Belongie, James Hays, Pietro Perona, Deva Ramanan, Piotr Doll{\'a}r, and C~Lawrence Zitnick.
\newblock Microsoft coco: Common objects in context.
\newblock In \emph{ECCV}, 2014.

\bibitem[Lin et~al.(2020)Lin, Ding, Zeng, and Tao]{lin2020gps}
Xin Lin, Changxing Ding, Jinquan Zeng, and Dacheng Tao.
\newblock Gps-net: Graph property sensing network for scene graph generation.
\newblock In \emph{CVPR}, 2020.

\bibitem[Liu et~al.(2023)Liu, Tang, Ni, Lu, Zhao, Li, and Tan]{liu2023large}
Xuejing Liu, Wei Tang, Xinzhe Ni, Jinghui Lu, Rui Zhao, Zechao Li, and Fei Tan.
\newblock What large language models bring to text-rich vqa?
\newblock \emph{arXiv preprint arXiv:2311.07306}, 2023.

\bibitem[Loshchilov and Hutter(2019)]{loshchilov2017decoupled}
Ilya Loshchilov and Frank Hutter.
\newblock Decoupled weight decay regularization.
\newblock In \emph{ICLR}, 2019.

\bibitem[Lu et~al.(2016)Lu, Krishna, Bernstein, and Fei-Fei]{lu2016visual}
Cewu Lu, Ranjay Krishna, Michael Bernstein, and Li~Fei-Fei.
\newblock Visual relationship detection with language priors.
\newblock In \emph{ECCV}, 2016.

\bibitem[Min et~al.(2023)Min, Ross, Sulem, Veyseh, Nguyen, Sainz, Agirre, Heintz, and Roth]{min2023recent}
Bonan Min, Hayley Ross, Elior Sulem, Amir Pouran~Ben Veyseh, Thien~Huu Nguyen, Oscar Sainz, Eneko Agirre, Ilana Heintz, and Dan Roth.
\newblock Recent advances in natural language processing via large pre-trained language models: A survey.
\newblock \emph{ACM Computing Surveys}, pages 1--40, 2023.

\bibitem[OpenAI(2022)]{chatgpt}
OpenAI.
\newblock Chatgpt, 2022.
\newblock URL \url{https://chat.openai.com/}.

\bibitem[Peyre et~al.(2017)Peyre, Sivic, Laptev, and Schmid]{peyre2017weakly}
Julia Peyre, Josef Sivic, Ivan Laptev, and Cordelia Schmid.
\newblock Weakly-supervised learning of visual relations.
\newblock In \emph{ICCV}, 2017.

\bibitem[Qi et~al.(2019)Qi, Li, Yang, Wang, and Luo]{qi2019attentive}
Mengshi Qi, Weijian Li, Zhengyuan Yang, Yunhong Wang, and Jiebo Luo.
\newblock Attentive relational networks for mapping images to scene graphs.
\newblock In \emph{CVPR}, 2019.

\bibitem[Shah et~al.(2023)Shah, Osi{\'n}ski, Levine, et~al.]{shah2023lm}
Dhruv Shah, B{\l}a{\.z}ej Osi{\'n}ski, Sergey Levine, et~al.
\newblock Lm-nav: Robotic navigation with large pre-trained models of language, vision, and action.
\newblock In \emph{Conference on Robot Learning}, pages 492--504, 2023.

\bibitem[Shi et~al.(2019)Shi, Zhang, and Li]{shi2019explainable}
Jiaxin Shi, Hanwang Zhang, and Juanzi Li.
\newblock Explainable and explicit visual reasoning over scene graphs.
\newblock In \emph{CVPR}, 2019.

\bibitem[Shit et~al.(2022)Shit, Koner, Wittmann, Paetzold, Ezhov, Li, Pan, Sharifzadeh, Kaissis, Tresp, et~al.]{shit2022relationformer}
Suprosanna Shit, Rajat Koner, Bastian Wittmann, Johannes Paetzold, Ivan Ezhov, Hongwei Li, Jiazhen Pan, Sahand Sharifzadeh, Georgios Kaissis, Volker Tresp, et~al.
\newblock Relationformer: A unified framework for image-to-graph generation.
\newblock In \emph{ECCV}, 2022.

\bibitem[Singh et~al.(2023)Singh, Salvador, Weihs, and Kembhavi]{singh2023scene}
Kunal~Pratap Singh, Jordi Salvador, Luca Weihs, and Aniruddha Kembhavi.
\newblock Scene graph contrastive learning for embodied navigation.
\newblock In \emph{ICCV}, 2023.

\bibitem[Su et~al.(2022)Su, Zhu, Murtadha, Pan, Wen, and Liu]{su2022zlpr}
Jianlin Su, Mingren Zhu, Ahmed Murtadha, Shengfeng Pan, Bo~Wen, and Yunfeng Liu.
\newblock Zlpr: A novel loss for multi-label classification.
\newblock \emph{arXiv preprint arXiv:2208.02955}, 2022.

\bibitem[Tang et~al.(2023)Tang, Zheng, Yu, and Yang]{tang2023cotdet}
Jiajin Tang, Ge~Zheng, Jingyi Yu, and Sibei Yang.
\newblock Cotdet: Affordance knowledge prompting for task driven object detection.
\newblock In \emph{ICCV}, 2023.

\bibitem[Tang et~al.(2019)Tang, Zhang, Wu, Luo, and Liu]{tang2019learning}
Kaihua Tang, Hanwang Zhang, Baoyuan Wu, Wenhan Luo, and Wei Liu.
\newblock Learning to compose dynamic tree structures for visual contexts.
\newblock In \emph{CVPR}, 2019.

\bibitem[Tang et~al.(2020)Tang, Niu, Huang, Shi, and Zhang]{tang2020unbiased}
Kaihua Tang, Yulei Niu, Jianqiang Huang, Jiaxin Shi, and Hanwang Zhang.
\newblock Unbiased scene graph generation from biased training.
\newblock In \emph{CVPR}, 2020.

\bibitem[Touvron et~al.(2023{\natexlab{a}})Touvron, Lavril, Izacard, Martinet, Lachaux, Lacroix, Rozi{\`e}re, Goyal, Hambro, Azhar, et~al.]{touvron2023llama1}
Hugo Touvron, Thibaut Lavril, Gautier Izacard, Xavier Martinet, Marie-Anne Lachaux, Timoth{\'e}e Lacroix, Baptiste Rozi{\`e}re, Naman Goyal, Eric Hambro, Faisal Azhar, et~al.
\newblock Llama: Open and efficient foundation language models.
\newblock \emph{arXiv preprint arXiv:2302.13971}, 2023{\natexlab{a}}.

\bibitem[Touvron et~al.(2023{\natexlab{b}})Touvron, Martin, Stone, Albert, Almahairi, Babaei, Bashlykov, Batra, Bhargava, Bhosale, et~al.]{touvron2023llama}
Hugo Touvron, Louis Martin, Kevin Stone, Peter Albert, Amjad Almahairi, Yasmine Babaei, Nikolay Bashlykov, Soumya Batra, Prajjwal Bhargava, Shruti Bhosale, et~al.
\newblock Llama 2: Open foundation and fine-tuned chat models.
\newblock \emph{arXiv preprint arXiv:2307.09288}, 2023{\natexlab{b}}.

\bibitem[Tsai et~al.(2023)Tsai, Dhar, Li, Zhang, and Zhang]{tsai2023multimodal}
Yao-Hung~Hubert Tsai, Vansh Dhar, Jialu Li, Bowen Zhang, and Jian Zhang.
\newblock Multimodal large language model for visual navigation.
\newblock \emph{arXiv preprint arXiv:2310.08669}, 2023.

\bibitem[Vaswani et~al.(2017)Vaswani, Shazeer, Parmar, Uszkoreit, Jones, Gomez, Kaiser, and Polosukhin]{vaswani2017attention}
Ashish Vaswani, Noam Shazeer, Niki Parmar, Jakob Uszkoreit, Llion Jones, Aidan~N Gomez, {\L}ukasz Kaiser, and Illia Polosukhin.
\newblock Attention is all you need.
\newblock \emph{NeurIPS}, 2017.

\bibitem[Wang et~al.(2023{\natexlab{a}})Wang, Wen, Li, Guo, Yang, and Liu]{wang2023pair}
Jinghao Wang, Zhengyu Wen, Xiangtai Li, Zujin Guo, Jingkang Yang, and Ziwei Liu.
\newblock Pair then relation: Pair-net for panoptic scene graph generation, 2023{\natexlab{a}}.

\bibitem[Wang et~al.(2023{\natexlab{b}})Wang, Shi, Li, Wang, Huang, Xing, Chen, Li, Zhu, Cao, et~al.]{wang2023all}
Weiyun Wang, Min Shi, Qingyun Li, Wenhai Wang, Zhenhang Huang, Linjie Xing, Zhe Chen, Hao Li, Xizhou Zhu, Zhiguo Cao, et~al.
\newblock The all-seeing project: Towards panoptic visual recognition and understanding of the open world.
\newblock \emph{arXiv preprint arXiv:2308.01907}, 2023{\natexlab{b}}.

\bibitem[Wei et~al.(2022)Wei, Wang, Schuurmans, Bosma, Xia, Chi, Le, Zhou, et~al.]{wei2022chain}
Jason Wei, Xuezhi Wang, Dale Schuurmans, Maarten Bosma, Fei Xia, Ed~Chi, Quoc~V Le, Denny Zhou, et~al.
\newblock Chain-of-thought prompting elicits reasoning in large language models.
\newblock \emph{NeurIPS}, 2022.

\bibitem[Xenos et~al.(2023)Xenos, Stafylakis, Patras, and Tzimiropoulos]{xenos2023simple}
Alexandros Xenos, Themos Stafylakis, Ioannis Patras, and Georgios Tzimiropoulos.
\newblock A simple baseline for knowledge-based visual question answering.
\newblock \emph{arXiv preprint arXiv:2310.13570}, 2023.

\bibitem[Xu et~al.(2017)Xu, Zhu, Choy, and Fei-Fei]{xu2017scene}
Danfei Xu, Yuke Zhu, Christopher~B Choy, and Li~Fei-Fei.
\newblock Scene graph generation by iterative message passing.
\newblock In \emph{CVPR}, 2017.

\bibitem[Yang et~al.(2023)Yang, Zhang, Li, Zou, Li, and Gao]{yang2023setofmark}
Jianwei Yang, Hao Zhang, Feng Li, Xueyan Zou, Chunyuan Li, and Jianfeng Gao.
\newblock Set-of-mark prompting unleashes extraordinary visual grounding in gpt-4v.
\newblock \emph{arXiv preprint arXiv:2310.11441}, 2023.

\bibitem[Yang et~al.(2022)Yang, Ang, Guo, Zhou, Zhang, and Liu]{yang2022panoptic}
Jingkang Yang, Yi~Zhe Ang, Zujin Guo, Kaiyang Zhou, Wayne Zhang, and Ziwei Liu.
\newblock Panoptic scene graph generation.
\newblock In \emph{ECCV}, 2022.

\bibitem[Yu et~al.(2021)Yu, Chai, Wang, Hu, and Wu]{yu2020cogtree}
Jing Yu, Yuan Chai, Yujing Wang, Yue Hu, and Qi~Wu.
\newblock Cogtree: Cognition tree loss for unbiased scene graph generation.
\newblock In \emph{IJCAI}, 2021.

\bibitem[Yu et~al.(2023)Yu, Li, Wu, Tang, Ji, and Zhuang]{yu2023visually}
Qifan Yu, Juncheng Li, Yu~Wu, Siliang Tang, Wei Ji, and Yueting Zhuang.
\newblock Visually-prompted language model for fine-grained scene graph generation in an open world.
\newblock \emph{arXiv preprint arXiv:2303.13233}, 2023.

\bibitem[Zareian et~al.(2020)Zareian, Karaman, and Chang]{zareian2020bridging}
Alireza Zareian, Svebor Karaman, and Shih-Fu Chang.
\newblock Bridging knowledge graphs to generate scene graphs.
\newblock In \emph{ECCV}, 2020.

\bibitem[Zellers et~al.(2018)Zellers, Yatskar, Thomson, and Choi]{zellers2018neural}
Rowan Zellers, Mark Yatskar, Sam Thomson, and Yejin Choi.
\newblock Neural motifs: Scene graph parsing with global context.
\newblock In \emph{CVPR}, 2018.

\bibitem[Zhang et~al.(2022)Zhang, Yao, Chen, Ji, Liu, Sun, and Chua]{zhang2022fine}
Ao~Zhang, Yuan Yao, Qianyu Chen, Wei Ji, Zhiyuan Liu, Maosong Sun, and Tat-Seng Chua.
\newblock Fine-grained scene graph generation with data transfer.
\newblock In \emph{ECCV}, 2022.

\bibitem[Zhang et~al.(2023)Zhang, Zhao, Xie, Zheng, Ji, and Chua]{zhang2023next}
Ao~Zhang, Liming Zhao, Chen-Wei Xie, Yun Zheng, Wei Ji, and Tat-Seng Chua.
\newblock Next-chat: An lmm for chat, detection and segmentation.
\newblock \emph{arXiv preprint arXiv:2311.04498}, 2023.

\bibitem[Zhang et~al.(2017)Zhang, Kyaw, Chang, and Chua]{zhang2017visual}
Hanwang Zhang, Zawlin Kyaw, Shih-Fu Chang, and Tat-Seng Chua.
\newblock Visual translation embedding network for visual relation detection.
\newblock In \emph{CVPR}, 2017.

\bibitem[Zhang et~al.(2019{\natexlab{a}})Zhang, Kalantidis, Rohrbach, Paluri, Elgammal, and Elhoseiny]{zhang2019large}
Ji~Zhang, Yannis Kalantidis, Marcus Rohrbach, Manohar Paluri, Ahmed Elgammal, and Mohamed Elhoseiny.
\newblock Large-scale visual relationship understanding.
\newblock In \emph{AAAI}, 2019{\natexlab{a}}.

\bibitem[Zhang et~al.(2019{\natexlab{b}})Zhang, Shih, Elgammal, Tao, and Catanzaro]{zhang2019graphical}
Ji~Zhang, Kevin~J Shih, Ahmed Elgammal, Andrew Tao, and Bryan Catanzaro.
\newblock Graphical contrastive losses for scene graph parsing.
\newblock In \emph{CVPR}, 2019{\natexlab{b}}.

\bibitem[Zheng et~al.(2019)Zheng, Chen, and Jin]{zheng2019visual}
Sipeng Zheng, Shizhe Chen, and Qin Jin.
\newblock Visual relation detection with multi-level attention.
\newblock In \emph{ACM MM}, 2019.

\bibitem[Zhou et~al.(2023{\natexlab{a}})Zhou, Alabi, Wei, Vercauteren, and Shi]{zhou2023text}
Zijian Zhou, Oluwatosin Alabi, Meng Wei, Tom Vercauteren, and Miaojing Shi.
\newblock Text promptable surgical instrument segmentation with vision-language models.
\newblock \emph{arXiv preprint arXiv:2306.09244}, 2023{\natexlab{a}}.

\bibitem[Zhou et~al.(2023{\natexlab{b}})Zhou, Shi, and Caesar]{zhou2023hilo}
Zijian Zhou, Miaojing Shi, and Holger Caesar.
\newblock Hilo: Exploiting high low frequency relations for unbiased panoptic scene graph generation.
\newblock In \emph{ICCV}, 2023{\natexlab{b}}.

\end{thebibliography}
}


\newpage

\section*{Appendix}
\appendix

\noindent In this supplementary material, we provide more details about our experiments,  additional experimental results, the RP- and RJ-prompts along with the corresponding descriptions output by the LLMs, as well as further visualization results.
First we provide more details about our experiments in Sec.~\ref{sec:more_details}
Second, we provide additional experimental results in Sec.~\ref{sec:more_exp}.
Then Sec.~\ref{sec:prompt} presents specific examples of RP-prompts and RJ-prompts applied to two LLMs, GPT-3.5 Turbo~\cite{chatgpt} and Llama2-7B~\cite{touvron2023llama}.
Sec.~\ref{sec:description} provides descriptions given by GPT-3.5 Turbo and Llama2-7B for various subjects, objects, and relations.
Sec.~\ref{sec:analysis_llm} offers a comparative analysis of descriptions from these different language models.
Finally, we provide more visualizations to illustrate the performance of our VLPrompt in Sec.~\ref{sec:more_vis}.

\section{More details about our experiments}
\label{sec:more_details}

\subsection{Tasks and metrics}

\noindent \textbf{Tasks}.
There are three subtasks for PSG and SGG tasks: Predicate Classification, Scene Graph Classification and Scene Graph Detection~\cite{xu2017scene}.
We focus on Scene Graph Detection for both datasets, as it is the most challenging and comprehensive subtask, which involves localizing objects and predicting their classes and relations.

\noindent \textbf{Metrics}.
Following previous works~\cite{yang2022panoptic, zhou2023hilo, wang2023pair}, we use Recall@K (R@K) and mean Recall@K (mR@K) as our metrics.
While Recall@K is biased towards frequent classes, mean Recall@K gives all classes the same weight.

\subsection{Implementation details}

In our experiments, we use Mask2Former~\cite{cheng2022masked} pretrained on COCO~\cite{lin2014microsoft} dataset to initialize the panoptic segmentation network in the vision feature extractor.
In language feature extractor, we utilize by default GPT-3.5 Turbo~\cite{chatgpt} as the LLM, and employ OpenAI Embedding Service~\cite{chatgpt} as the text encoder.
We store the extracted language prompting features in a database and then retrieve the RP-language prompting feature using ``sub\#obj'', and the RJ-language prompting feature using ``sub\#rel\#obj''.
We adopt the same data augmentation settings following previous methods~\cite{yang2022panoptic, zhou2023hilo}.
To train our model, we use the AdamW~\cite{loshchilov2017decoupled}, with a learning rate of $1e^{-4}$ and weight decay of $5e^{-2}$.
We set $\lambda$ to 0.1 in our final loss function.
Our model is trained for 12 epochs with a step scheduler reducing the learning rate to $1e^{-5}$ at epoch 6 and further to $1e^{-6}$ at epoch 10.
The training takes approximately 18 hours on four A100 GPUs, with a batch size of 1 for each GPU.
The inference of our model follows the same forward process in training.

\section{More experimental results}
\label{sec:more_exp}


\noindent \textbf{Different LLMs and text encoders.}
To further assess the effects of different LLMs and text encoders on model performance, we attempt to replace the GPT with Llama2-7B~\cite{touvron2023llama} and the OpenAI Embedding Service with Bert \cite{devlin2018bert}.
When using Bert, we take the mean of the embeddings of all output tokens as the feature.
The experimental results (Llama2-7B + Bert) in Tab.~\ref{tab:more_exp} reveal that using Llama2 as the LLM and Bert as the text encoder results in only a slight decrease in performance: 0.4\% in R@100 and 0.5\% in mR@100.
We review the descriptions output by Llama2-7B and compare them with those from GPT-3.5 Turbo, finding no significant differences in quality, more details can be found in supplementary materials.
This suggests that, with carefully designed prompts, open-source LLMs with reduced parameters also work for our method, thus validating the flexibility of our method.

\noindent \textbf{Compress the language prompting feature.}
At runtime, the language prompting feature occupies 290.57MB of memory, which is manageable.
To further enhance the model's applicability in real-world scenarios, such as when there are more object and relation categories, we compress the language prompting feature to 1/4 of its original size using an encoder-decoder models.
Results (Compression) shown in Tab.~\ref{tab:more_exp} indicate that compressing the language prompting feature to 1/4 leads to only a minor performance decrease, while the memory required by the model during runtime is reduced to 1/4 of the original, which further demonstrates the practicality of our method in real-world settings.

\begin{table}[t]\small
    \centering
    \caption{More experimental results for our VLPrompt.}
    \resizebox{0.8\linewidth}{!}{
    \begin{tabular}{l|cc|cc|cc}
    \hline
        Method & R@20 & mR@20 & R@50 & mR@50 & R@100 & mR@100 \\
        \hline
        VLPrompt & 39.4 & 34.7 & 47.6 & 45.1 & 52.4 & 53.7 \\
        \hdashline
        Llama2 + Bert & 39.0 & 33.5 & 47.3 & 44.8 & 52.0 & 53.2 \\
        Compression & 38.3 & 33.1 & 46.5 & 44.6 & 51.2 & 51.1 \\
        \hline
    \end{tabular}
    }
    \label{tab:more_exp}
\end{table}

\noindent \textbf{Efficiency analysis.}
In Tab.~\ref{tab:analysis_efficiency}, we compare our model's efficiency with the previous state-of-the-art models, evaluating computational floating point operations per second (FLOPS), parameter size and inference speed on the same A100 GPU.
We observe that although our method has higher FLOPS compared to HiLo, it matches HiLo in prediction speed, and significantly outperforms HiLo in performance 
(see Tab.~1 of the paper).

\begin{table}[t]\small
    \centering
    \caption{Analysis for efficiency.}
    \vspace{-2mm}
    \resizebox{0.8\linewidth}{!}{
    \begin{tabular}{l|ccc}
    \hline
        Method & FLOPS (G) & Parameters (G) & Inference Speed (ms) \\
        \hline
        PSGTR \cite{yang2022panoptic} & 461.3 & 44.2 & 140 \\
        HiLo \cite{zhou2023hilo} & 229.4 & 58.7 & 156 \\
        VLPrompt (ours) & 386.7 & 49.1 & 152 \\
        \hline
    \end{tabular}
    }
    \vspace{-6mm}
    \label{tab:analysis_efficiency}
\end{table}

\section{RP- and RJ-prompt}
\label{sec:prompt}

According to the official API interfaces provided by GPT-3.5 Turbo and Llama2, we can interact with large language models (LLMs) in three different roles: ``system'', ``assistant'', and ``user''.
The system message defines the behavior of the assistant within a given context.
The assistant represents the responses and actions of the LLM, while the user refers to the individual engagement with the LLM, providing requests or comments for the assistant to address.
Inspired by the chain-of-thought~\cite{wei2022chain} technique, we have designed a meaningful dialogue as the input prompt for LLMs.
Specifically, the prompts designed for GPT-3.5 Turbo and Llama2-7B are presented below.

\subsection{Prompt for GPT-3.5 Turbo}
\label{sec:prompt_gpt}

\noindent \textbf{RP-prompt}.
Relation proposer prompt, which we used to stimulate LLMs to propose all possible relations between two given objects.

\underline{system}: \textit{You are asked to play the role of a} \textbf{relation proposer}\textit{. Given the category names of two objects in an image, you are to infer what kind of relation might exist between them based on your knowledge, and provide the reasons for each possible relation. In the relation between the two objects in the image, we refer to one object as the subject and the other as the object. There may or may not be a relation between the subject and the object. Please note that this relation has an order, that is, the subject comes first and the object comes after. If there is a relation between the two, these relations must belong to one of the pre-defined 56 different types.}

\underline{assistant}: \textit{What are the 56 relations?}

\underline{user}: \textit{They are `over', `in front of', `beside', `on', `in', `attached to', `hanging from', `on back of', `falling off', `going down', `painted on', `walking on', `running on', `crossing', `standing on', `lying on', `sitting on', `flying over', `jumping over', `jumping from', `wearing', `holding', `carrying', `looking at', `guiding', `kissing', `eating', `drinking', `feeding', `biting', `catching', `picking', `playing with', `chasing', `climbing', `cleaning', `playing', `touching', `pushing', `pulling', `opening', `cooking', `talking to', `throwing', `slicing', `driving', `riding', `parked on', `driving on', `about to hit', `kicking', `swinging', `entering', `exiting', `enclosing', `leaning on'.}

\underline{assistant}: \textit{Can you give me an example?}

\underline{user}: \textit{For example, the subject is a person, and the object is a sports ball. The possible relations between them could be: 1. Beside: The person could be standing beside the sports ball. 2. Looking at: The person might be looking at the ball to better control it. 3. Playing: This is because it's very common in real life for a person to be playing with a sports ball. 4. Chasing: The person might be chasing after the ball.}

\underline{assistant}: \textit{Ok, I got it. Please give me the subject and object of the image.}

\underline{user}: \textit{The subject is a SUBJECT\_NAME, and the object is a OBJECT\_NAME.}

\noindent \textbf{RJ-prompt}.
Relation judger prompt, which we used to stimulate LLMs to judge whether there is a specific relation between two objects.

\underline{system}: \textit{You are asked to play the role of a} \textbf{relation judger}\textit{. Given the category names of two objects in an image, and providing you with a relation category name, you need to predict whether this relation is likely to exist in the image based on your knowledge, and give the reason for its existence. For two objects, we call the first object subject and the second object object.}

\underline{assistant}: \textit{Yes, I understand. Can you give me an example?}

\underline{user}: \textit{For example, the input is: the subject is a 'person', the object is a 'sports ball' and the relation is 'playing'. The output should be Yes, the relation is likely to exist in the image. This is because it's very common in real life for a person to be playing with a sports ball.}

\underline{assistant}: \textit{Ok, I got it. Please give me the subject, object and relation names.}

\underline{user}: \textit{The subject is a SUBJECT\_NAME, the object is a OBJECT\_NAME, and the relation is RELATION\_NAME.}

\subsection{Prompt for Llama2-7B}
\label{sec:prompt_llama}
In GPT-3.5 Turbo, the ``system'' role can be followed by either the ``assistant'' or the ``user'' role.
However, in Llama2's API, the ``system'' role can only be followed by ``user'' role, not ``assistant'' role.
Therefore, we have made simple modifications to the prompt used for GPT-3.5 Turbo. The specific prompt is as follows.

\noindent \textbf{RP-prompt}.
Relation proposer prompt, which we used to stimulate LLMs to propose all possible relations between two given objects.

\underline{system}: \textit{You are asked to play the role of a} \textbf{relation proposer}\textit{. Given the category names of two objects in an image, you are to infer what kind of relation might exist between them based on your knowledge, and provide the reasons for each possible relation. In the relation between the two objects in the image, we refer to one object as the subject and the other as the object. There may or may not be a relation between the subject and the object. Please note that this relation has an order, that is, the subject comes first and the object comes after. If there is a relation between the two, these relations must belong to one of the pre-defined 56 different types. What are the 56 relations?}

\underline{user}: \textit{They are `over', `in front of', `beside', `on', `in', `attached to', `hanging from', `on back of', `falling off', `going down', `painted on', `walking on', `running on', `crossing', `standing on', `lying on', `sitting on', `flying over', `jumping over', `jumping from', `wearing', `holding', `carrying', `looking at', `guiding', `kissing', `eating', `drinking', `feeding', `biting', `catching', `picking', `playing with', `chasing', `climbing', `cleaning', `playing', `touching', `pushing', `pulling', `opening', `cooking', `talking to', `throwing', `slicing', `driving', `riding', `parked on', `driving on', `about to hit', `kicking', `swinging', `entering', `exiting', `enclosing', `leaning on'.}

\underline{assistant}: \textit{Can you give me an example?}

\underline{user}: \textit{For example, the subject is a person, and the object is a sports ball. The possible relations between them could be: 1. Beside: The person could be standing beside the sports ball. 2. Looking at: The person might be looking at the ball to better control it. 3. Playing: This is because it's very common in real life for a person to be playing with a sports ball. 4. Chasing: The person might be chasing after the ball.}

\underline{assistant}: \textit{Ok, I got it. Please give me the subject and object of the image.}

\underline{user}: \textit{The subject is a SUBJECT\_NAME, and the object is a OBJECT\_NAME.}

\noindent \textbf{RJ-prompt}.
Relation judger prompt, which we used to stimulate LLMs to judge whether there is a specific relation between two objects.

\underline{system}: \textit{You are asked to play the role of a} \textbf{relation judger}\textit{. Given the category names of two objects in an image, and providing you with a relation category name, you need to predict whether this relation is likely to exist in the image based on your knowledge, and give the reason for its existence. For two objects, we call the first object subject and the second object object. Please give me an example.}

\underline{user}: \textit{For example, the input is: the subject is a 'person', the object is a 'sports ball' and the relation is 'playing'. The output should be Yes, the relation is likely to exist in the image. This is because it's very common in real life for a person to be playing with a sports ball.}

\underline{assistant}: \textit{Ok, I got it. Please give me the subject, object and relation names.}

\underline{user}: \textit{The subject is a SUBJECT\_NAME, the object is a OBJECT\_NAME, and the relation is RELATION\_NAME.}

\section{RP- and RJ-description}
\label{sec:description}

To gain a more comprehensive understanding of the outputs from LLMs, we selected two sets of subject-object pairs for demonstration, namely ``person-motorcycle'', ``rock-horse''.
Here, the former term in each pair represents the category of the subject, and the latter the category of the object.
The RP-descriptions are as below.
For each pair, we randomly select two different relations to showcase their corresponding RJ-descriptions.
Note that in the examples shown below for RP-descriptions, the template sentences that were added to complete the descriptions for all predefined $K$ relations (\eg it is not likely for SUBJECT\_NAME and OBJECT\_NAME to have the relation RELATION\_NAME) have been omitted.

\subsection{Description from GPT-3.5 Turbo}

\noindent \textbf{RP-description}.
The description generated by LLMs based on the RP-prompt.

\underline{person-motorcycle}: \textit{Based on the given subject (person) and object (motorcycle), here are some possible relations between them: 1.} \textbf{Riding}\textit{: The person could be riding the motorcycle, indicating that they are in control and using it as a means of transportation. 2.} \textbf{Driving}\textit{: Similar to riding, the person could be driving the motorcycle, suggesting that they are operating it. 3.} \textbf{About to hit}\textit{: This relation implies that the person is in a dangerous situation where they are about to collide with the motorcycle, indicating a potential accident or collision. 4.} \textbf{Leaning on}\textit{: The person might be leaning on the motorcycle, using it as a support or resting against it. 5.} \textbf{Parked on}\textit{: The motorcycle could be parked on or near the person, suggesting that they are in close proximity to it but not actively using or interacting with it. 6.} \textbf{Driving on}\textit{: This relation implies that the person is driving another vehicle and the motorcycle is on the road, indicating that they are sharing the same space or path. 7.} \textbf{Looking at}\textit{: The person might be looking at the motorcycle, either out of interest or curiosity, or perhaps considering using it. 8.} \textbf{Touching}\textit{: The person could be touching or interacting with the motorcycle, such as examining it or making adjustments. 9.} \textbf{Holding}\textit{: The person might be holding onto the motorcycle, either for support or to prevent it from falling over. 10.} \textbf{Exiting}\textit{: This relation suggests that the person was previously on or inside the motorcycle and is now in the process of getting off or exiting it. These are just a few possible relations between a person and a motorcycle, and the actual relation would depend on the context and specific details of the image.}

\underline{rock-horse}: \textit{Based on the given subject (rock) and object (horse), here are some possible relations between a rock and a horse: 1.} \textbf{Beside}\textit{: The rock could be positioned beside the horse, indicating their proximity to each other. 2.} \textbf{On}\textit{: The horse could be standing on the rock, using it as a platform or a vantage point. 3.} \textbf{In front of}\textit{: The rock could be in front of the horse, blocking its path or serving as an obstacle. 4.} \textbf{Attached to}\textit{: The rock could be attached to the horse, possibly as a weight or burden. These are just a few possible relations between a rock and a horse. The actual relation would depend on the context and specific details of the image.}

\noindent \textbf{RJ-description}.
The description generated by LLMs based on the RJ-prompt.

\underline{person-riding-motorcycle}: \textit{Based on my knowledge, the relation riding is} \textbf{likely} \textit{to exist between person and motorcycle. This is because it is common for a person to ride a motorcycle.}

\underline{person-drinking-motorcycle}: \textit{No, the relation drinking is} \textbf{not likely} \textit{to exist between person and motorcycle. This is because it is not common for a person to be drinking while interacting with a motorcycle.}





\underline{rock-attached to-horse}: \textit{No, the relation attached to is} \textbf{not likely} \textit{to exist in the image. This is because rocks are typically not attached to horses in real life.}

\underline{rock-falling off-horse}: \textit{No, the relation falling off is} \textbf{not likely} \textit{to exist in the image. This is because rocks cannot fall off horses as they are inanimate objects and do not have the ability to interact with horses in that way.}

\subsection{Description from Llama2-7B}

\noindent \textbf{RP-description}.
The description generated by LLMs based on the RP-prompt.

\underline{person-motorcycle}: \textit{Great! Based on the category names you provided, here are some possible relations between the subject (person) and the object (motorcycle): 1.} \textbf{Riding}\textit{: The person could be riding the motorcycle. 2.} \textbf{Standing next to}\textit{: The person might be standing next to the motorcycle, perhaps admiring it or preparing to ride it. 3.} \textbf{Holding onto}\textit{: The person could be holding onto the motorcycle to steady themselves while riding it. 4.} \textbf{Leaning on}\textit{: The person might be leaning on the motorcycle while taking a break or resting. 5.} \textbf{Fixing}\textit{: The person could be fixing the motorcycle, perhaps making repairs or adjustments. 6.} \textbf{Parked near}\textit{: The person could be parked near the motorcycle, perhaps intending to ride it later. 7.} \textbf{Getting on}\textit{: The person could be getting on the motorcycle, perhaps after having just parked it. 8.} \textbf{Falling off}\textit{: The person might be falling off the motorcycle, perhaps after losing control or experiencing an accident. 9.} \textbf{Facing}\textit{: The person could be facing the motorcycle, perhaps to check on it or to prepare to ride it. 10.} \textbf{Behind}\textit{: The person could be behind the motorcycle, perhaps to push it or to provide support while riding. Please let me know if any of these relations are correct or if you would like me to provide more options.}

\underline{rock-horse}: \textit{Great! Based on the category names you provided, here are some possible relations between the subject (rock) and the object (horse): 1.} \textbf{Beside}\textit{: The rock could be beside the horse, perhaps as the horse is grazing or resting. 2.} \textbf{Looking at}\textit{: The horse might be looking at the rock, perhaps curious about it or interested in it for some reason. 3.} \textbf{Near}\textit{: The rock could be near the horse, perhaps in a field or meadow where the horse is roaming. 4.} \textbf{Riding}\textit{: The horse could be ridden over the rock, perhaps as a way to traverse a difficult terrain or to reach a destination. 5.} \textbf{Jumping}\textit{: The horse could be jumping over the rock, perhaps in a competition or as a way to clear an obstacle. 6.} \textbf{Kicking}\textit{: The horse could be kicking the rock, perhaps as a way to dislodge it or to mark its territory. 7.} \textbf{Standing on}\textit{: The horse could be standing on the rock, perhaps as a way to get a better view or to rest. 8.} \textbf{Falling off}\textit{: The horse could be falling off the rock, perhaps due to a misstep or a sudden movement. 9.} \textbf{Crossing}\textit{: The horse could be crossing the rock, perhaps as a way to get to the other side of a river or stream. 10.} \textbf{Pushing}\textit{: The horse could be pushing the rock, perhaps as a way to move it out of the way or to create a path. Please let me know if you have any other questions or if you would like me to generate more relations.}

\noindent \textbf{RJ-description}.
The description generated by LLMs based on the RJ-prompt.

\underline{person-riding-motorcycle}: \textit{Based on my knowledge, I would say that the relation riding is} \textbf{likely} \textit{to exist in the image. People often ride motorcycles as a mode of transportation or recreation, so it\'s a common and plausible relation in real life. Therefore, I would give a Yes answer to the question of whether the relation exists in the image.}

\underline{person-drinking-motorcycle}: \textit{Based on my knowledge, the relation drinking is} \textbf{unlikely} \textit{to exist in the image. The reason for this is that it is not common for a person to be drinking a motorcycle. Motorcycles are vehicles that are typically used for transportation, not for drinking. Therefore, the relation drinking does not make sense in this context.}

\underline{rock-attached to-horse}: \textit{Based on the information provided, I would say that the relation attached to is} \textbf{unlikely} \textit{to exist in the image. A rock and a horse are very different objects and do not have a natural connection or relationship with each other. Therefore, it is unlikely that they would be attached to each other in a real-world scenario.}

\underline{rock-falling off-horse}: \textit{Based on my knowledge, the relation falling off is} \textbf{unlikely} \textit{to exist in an image of a rock and a horse. The reason for this is that rocks and horses are not typically associated with each other in a falling off context. Rocks are inanimate objects that do not have the ability to fall off, while horses are living creatures that are not known to fall off of other objects. Therefore, the probability of seeing a relation of falling off between a rock and a horse in an image is low.}

\section{Description analysis}
\label{sec:analysis_llm}

Through the series of examples above, we analyze the descriptions generated by the LLMs as follows.

\noindent \textbf{Comparation between GPT-3.5 Turbo and Llama2-7B}.
Comparing the descriptions generated by GPT-3.5 Turbo and Llama2-7B, we find that GPT-3.5 Turbo's descriptions are slightly better than those of Llama2-7B.
However, thanks to our carefully designed prompts, even using the Llama2-7B model can still produce high-quality descriptions to aid in relation prediction.
As shown in 
Tab.~4
of our paper, models trained using descriptions from Llama2-7B are only 0.5\% lower in mR@100 compared to those using GPT-3.5 Turbo.



\section{More visualizations}
\label{sec:more_vis}

\begin{figure*}[h]
    \centering
    \includegraphics[width=0.95\linewidth]{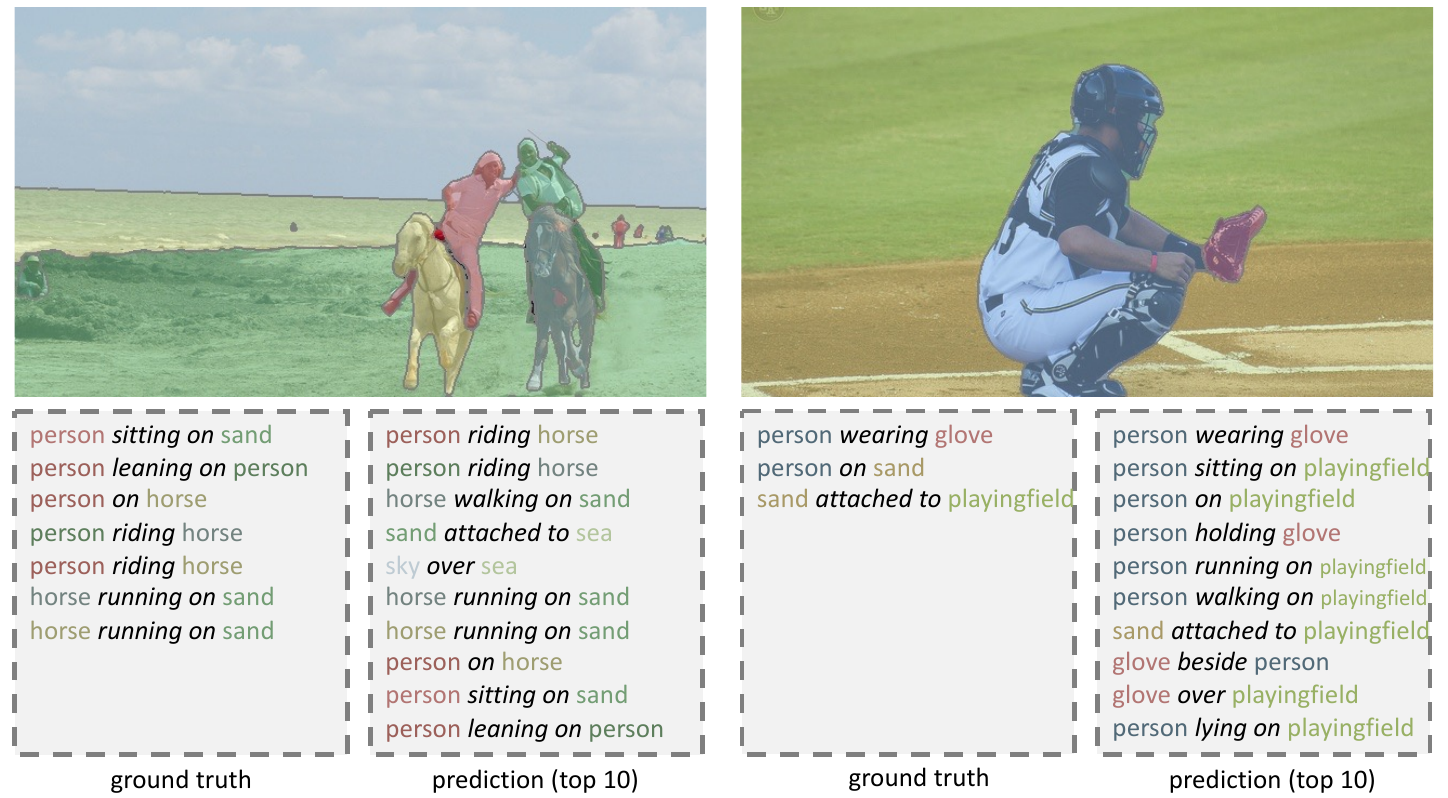}
    \caption{Visualization results of our VLPrompt.
    }
    \label{fig:vis_v2}
\end{figure*}

\begin{figure*}[h]
    \centering
    \includegraphics[width=0.95\linewidth]{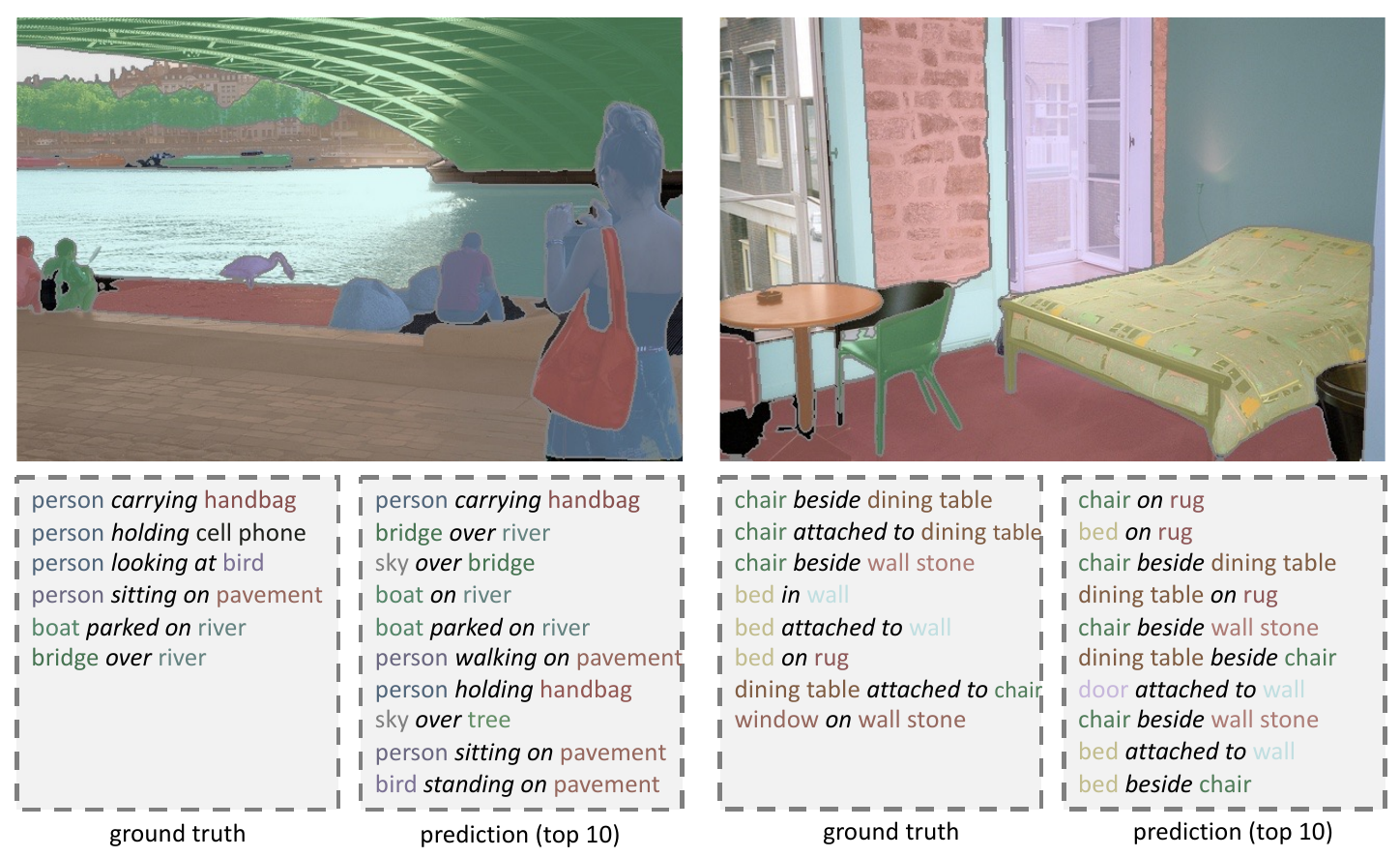}
    \caption{Visualization results of our VLPrompt.
    }
    \label{fig:vis_v3}
\end{figure*}

\begin{figure*}[h]
    \centering
    \includegraphics[width=0.95\linewidth]{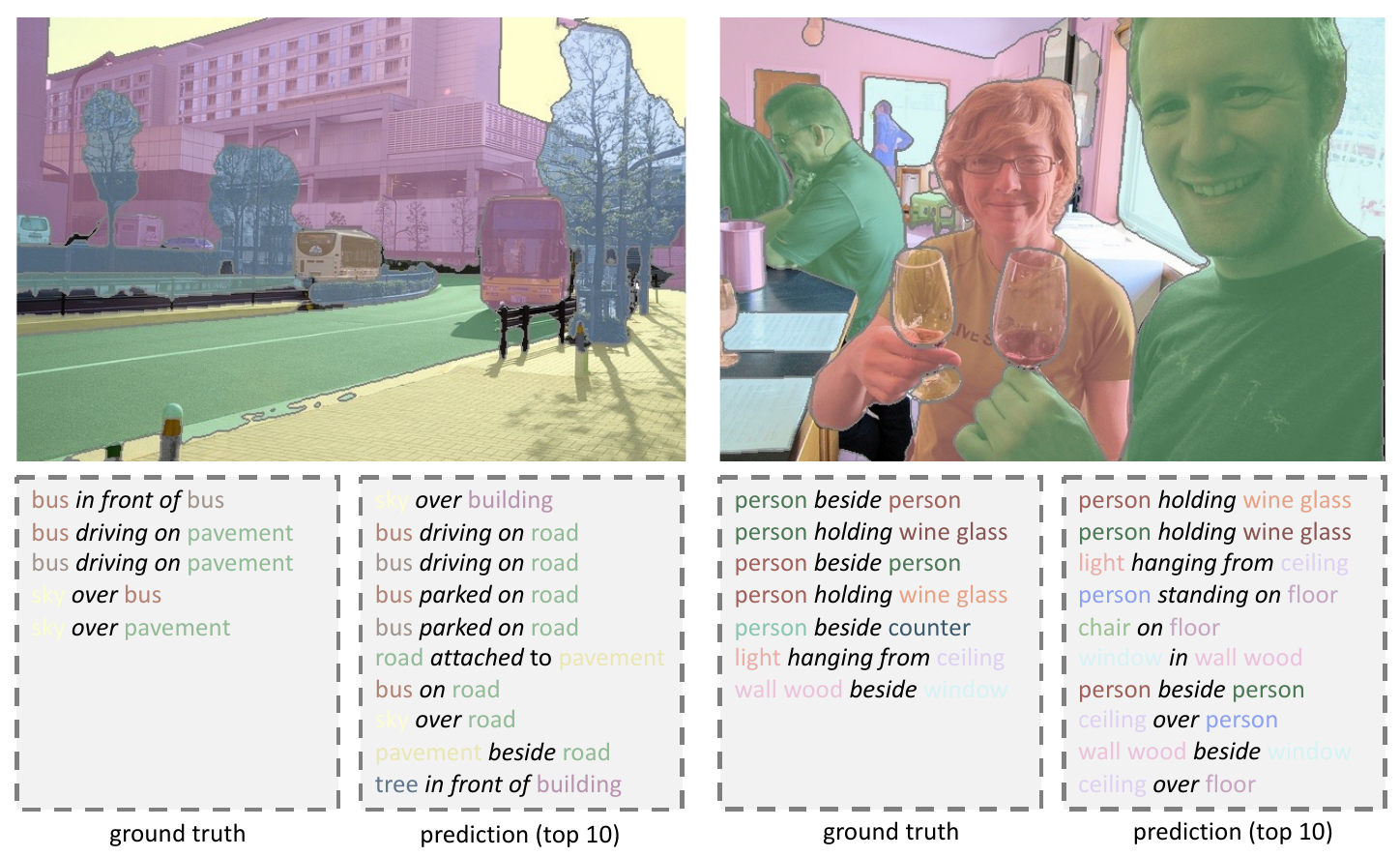}
    \caption{Visualization results of our VLPrompt.
    }
    \label{fig:vis_v4}
\end{figure*}

\begin{figure*}[h]
    \centering
    \includegraphics[width=0.95\linewidth]{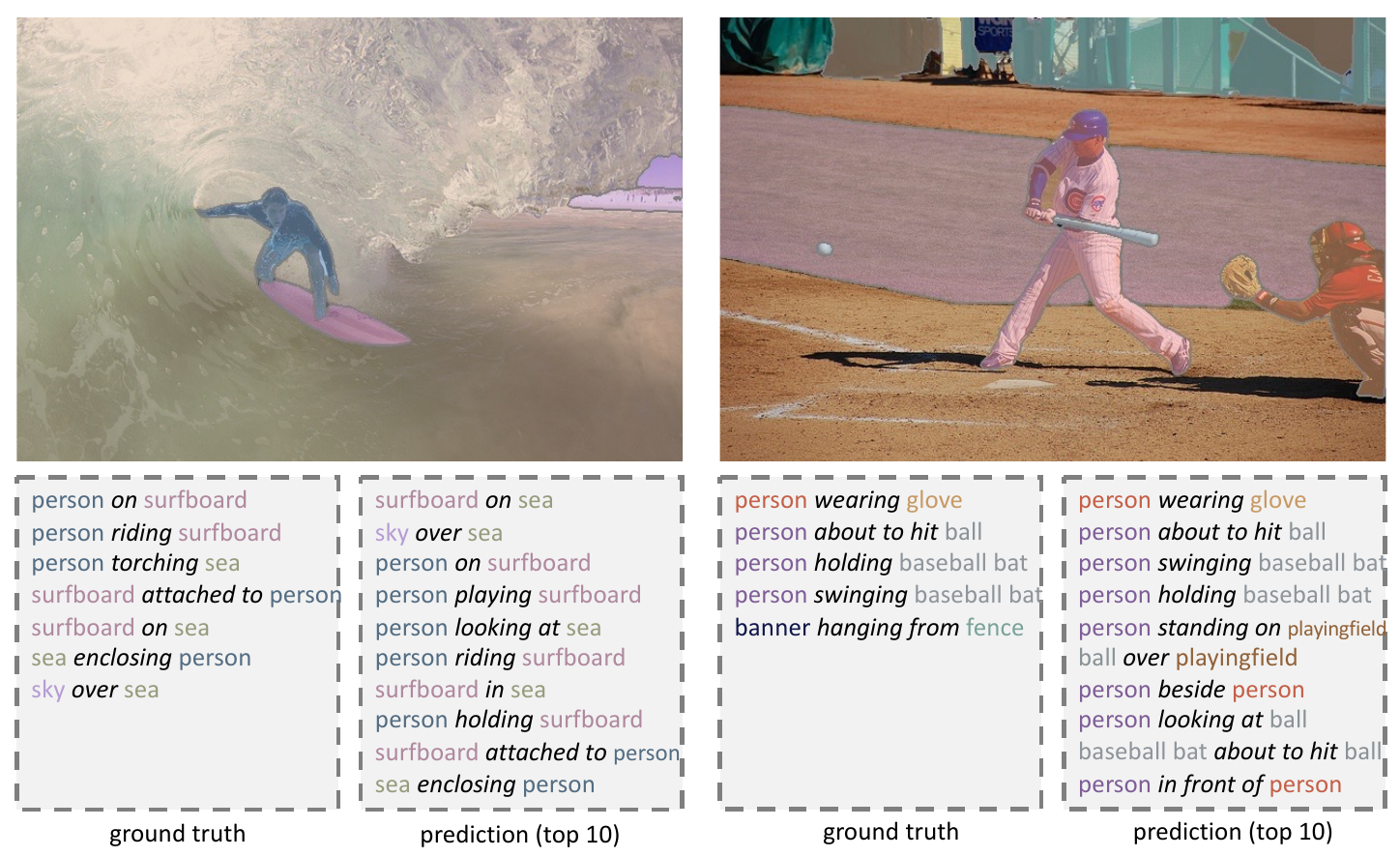}
    \caption{Visualization results of our VLPrompt.
    }
    \label{fig:vis_v5}
\end{figure*}

\begin{figure*}[h]
    \centering
    \includegraphics[width=0.95\linewidth]{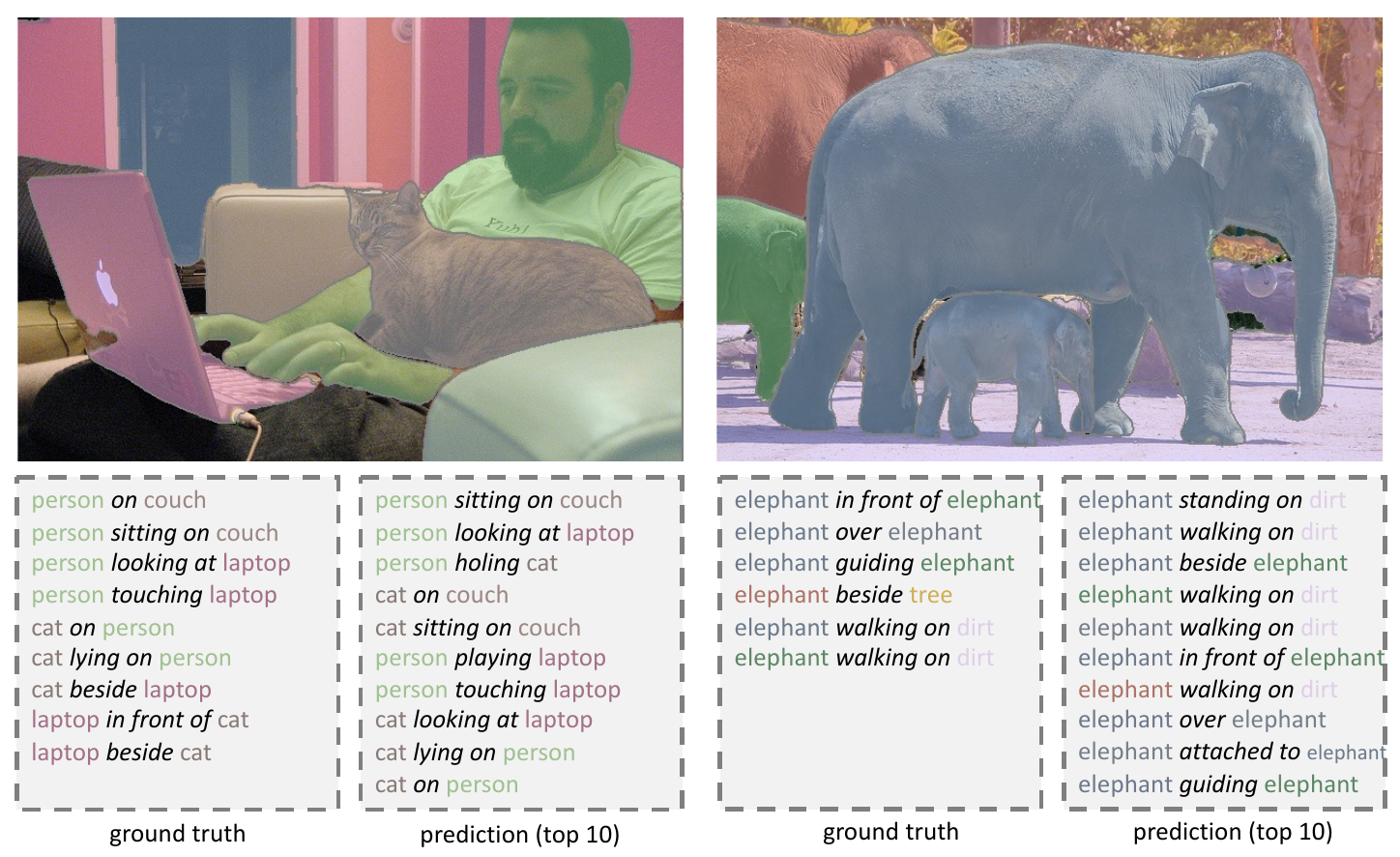}
    \caption{Visualization results of our VLPrompt.
    }
    \label{fig:vis_v6}
\end{figure*}

\begin{figure*}[h]
    \centering
    \includegraphics[width=0.95\linewidth]{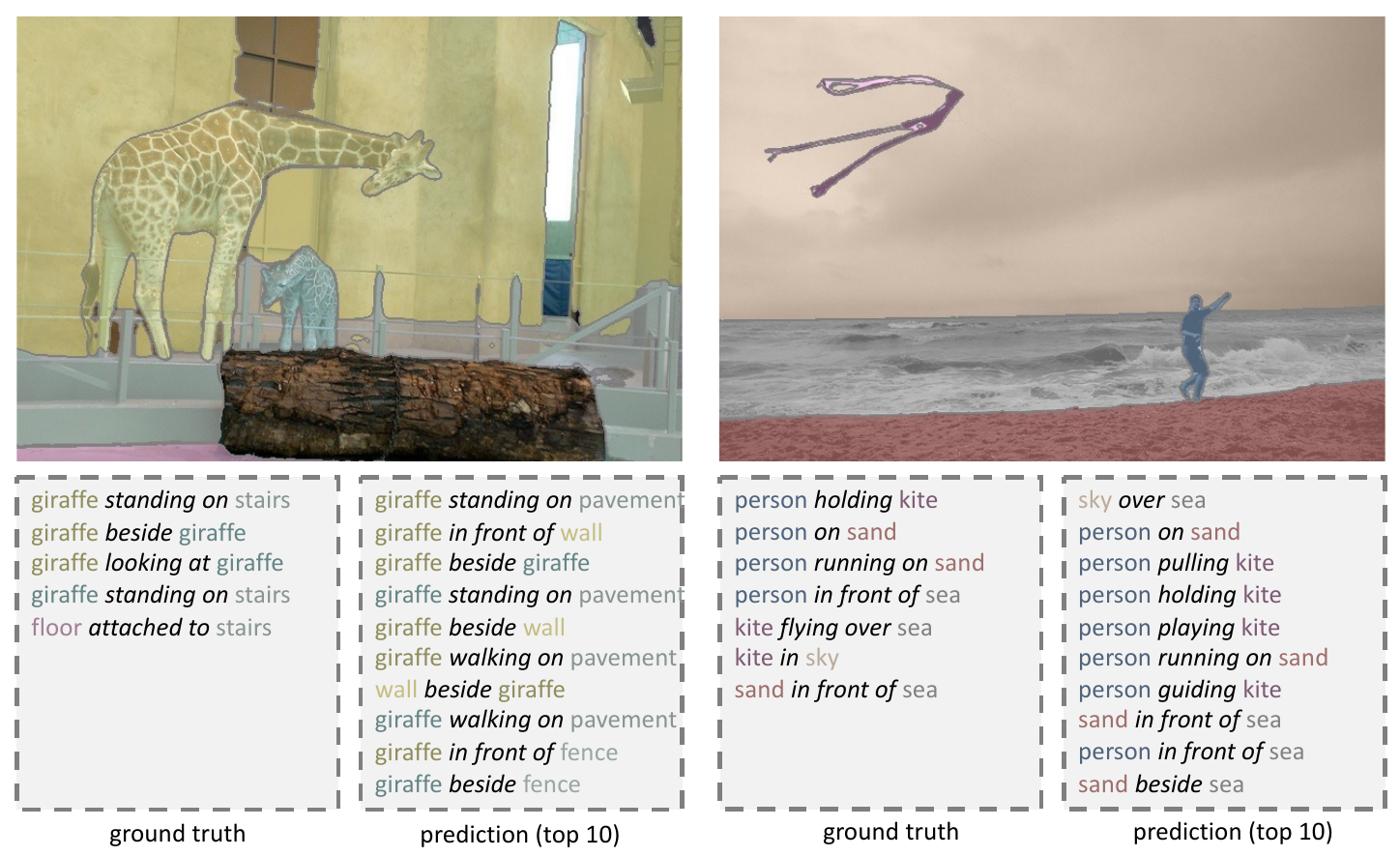}
    \caption{Visualization results of our VLPrompt.
    }
    \label{fig:vis_v7}
\end{figure*}

Fig.~\ref{fig:vis_v2} to \ref{fig:vis_v7} showcase additional visualization results of our VLPrompt.
Each figure contains two examples.
For each example, the top shows the results of panoptic segmentation, the bottom left displays the ground truth, and the bottom right shows the top 10 relation prediction results.
Based on visualization results, our VLPrompt exhibits precise capabilities in relation prediction, thereby enhancing scene understanding.
For instance, in Fig.~\ref{fig:vis_v2} on the left side, the top 10 relation predictions not only encompass all the relations from the ground truth but also include additional correct relations involving stuff categories.
However, we have also identified some bad cases, of which many are due to inaccurate panoptic segmentation predictions (including category classification and mask prediction).

\end{document}